\documentclass[10pt,journal,compsoc]{IEEEtran}
\ifCLASSOPTIONcompsoc
  \usepackage[nocompress]{cite}
\else
  \usepackage{cite}
\fi
\usepackage[normalem]{ulem}
\usepackage{multirow}
\usepackage{multicol}
\usepackage{graphicx}
\usepackage{booktabs}
\usepackage{orcidlink}
\usepackage{pifont}
\usepackage{color}
\usepackage{amsmath,amssymb}
 \usepackage[caption=true,font=footnotesize,labelfont=sf,textfont=sf]{subfig}
\newcommand{\cmark}{\ding{51}}%
\newcommand{\xmark}{\text{\ding{55}}}
\newcommand{\eg}{e.g.}

\begin{document}
\title{FlowFormer: A Transformer Architecture and Its Masked Cost Volume Autoencoding for Optical Flow}
%
%
% author names and IEEE mem2berships

\author{Zhaoyang Huang*,
        Xiaoyu Shi*, Chao Zhang, Qiang Wang, Yijin Li, Hongwei Qin, \\Jifeng Dai, Xiaogang Wang, and Hongsheng Li
\IEEEcompsocitemizethanks{
\IEEEcompsocthanksitem 
Zhaoyang Huang and Xiaoyu Shi assert equal contributions,
\IEEEcompsocthanksitem 
Zhaoyang Huang, Xiaoyu Shi, Xiaogang Wang, and Hongsheng Li are with the Department of Electronic Engineering, The Chinese University of Hong Kong, Hong Kong SAR. E-mail: \{drinkingcoder@link, xiaoyushi@link, xgwang@ee, hsli@ee\}.cuhk.edu.hk, 
\IEEEcompsocthanksitem
Chao Zhang and Qiang Wang are with Samsung Telecommunication Research 
\IEEEcompsocthanksitem
Yijin Li is with the State Key Lab of CAD$\&$CG in Zhejiang University.
\IEEEcompsocthanksitem
Hongwei Qin is with SenseTime Research, Shenzhen,
China. 
% Email: qinhwi@gmail.com
\IEEEcompsocthanksitem Jifeng Dai is with the Department of Electronic Engineering, Tsinghua University. Email: daijifeng@tsinghua.edu.cn
% \IEEEcompsocthanksitem 
% Chao Zhang and Qiang Wang are with SamSung
\protect
}
}


\markboth{Journal of \LaTeX\ Class Files,~Vol.~14, No.~8, August~2015}%
{Shell \MakeLowercase{\textit{et al.}}: Bare Demo of IEEEtran.cls for Computer Society Journals}

% \IEEEpubid{0000--0000/00\$00.00~\copyright~2015 IEEE}


\IEEEtitleabstractindextext{%

\begin{abstract}
This paper introduces a novel transformer-based network architecture, FlowFormer, along with the Masked Cost Volume AutoEncoding (MCVA) for pretraining it to tackle the problem of optical flow estimation.
FlowFormer tokenizes the 4D cost-volume built from the source-target image pair and iteratively refines flow estimation with a cost-volume encoder-decoder architecture.
The cost-volume encoder derives a cost memory with alternate-group transformer~(AGT) layers in a latent space and the decoder recurrently decodes flow from the cost memory with dynamic positional cost queries.
On the Sintel benchmark, FlowFormer architecture achieves 1.16 and 2.09 average end-point-error~(AEPE) on the clean and final pass, a 16.5\% and 15.5\% error reduction from the GMA~(1.388 and 2.47).
MCVA enhances FlowFormer by pretraining the cost-volume encoder with a masked autoencoding scheme, which further unleashes the capability of FlowFormer with unlabeled data.
This is especially critical in optical flow estimation because ground truth flows are more expensive to acquire than labels in other vision tasks.
MCVA improves FlowFormer all-sided and FlowFormer+MCVA ranks 1st among all published methods on both Sintel and KITTI-2015 benchmarks and achieves the best generalization performance. Specifically, FlowFormer+MCVA achieves 1.07 and 1.94 AEPE on the Sintel benchmark, leading to 7.76\% and 7.18\% error reductions from FlowFormer. 
%MCVA-FlowFormer obtains 4.52 F1-all on the KITTI-2015 test set, improving FlowFormer by 0.16.
% Besides, MCVA-FlowFormer also achieves strong generalization performance.
% Without being trained on Sintel, MCVA-FlowFormer achieves 0.90 AEPE on the clean and pass of Sintel training set, outperforming the best published result~(1.08) by 16.7\%.
% Pretraining a transformer-based neural network architecture with masked autoencoding (MAE) achieves great success in natural language processing and image recognition. 
% This paradigm unleashes the capability of transformers with unlabeled data.
% Optical flow, where the dense groundtruth is also hard to obtain, is a classical task in computer vision. 
% In this paper, we introduce the MAE with transformers to optical flow.
% Firstly, we propose a optical Flow transFormer, dubbed as FlowFormer, a transformer-based neural network architecture for learning
% optical flow,
% including a pretrained image feature encoder, a cost volume encoder, and a recurrent cost memory decoder.
% FlowFormer achieves state-of-the-art performance on both KITTI and Sintel benchmarks and shows the strongest generalization performance.
% To further unleash the capacity of the transformer-based cost volume encoder, we propose a novel masked cost volume autoencoding (MCAE) 
\end{abstract}
\begin{IEEEkeywords}
Optical Flow, Masked Autoencoding, Transformer.
\end{IEEEkeywords}}

\maketitle

\IEEEdisplaynontitleabstractindextext

% \fi
\IEEEpeerreviewmaketitle

\IEEEraisesectionheading{\section{Introduction}\label{sec:introduction}}
\begin{figure*}[!t]
    \begin{center}
    \includegraphics[width=0.95\linewidth, trim={20mm 135mm 260mm 90mm}, clip]{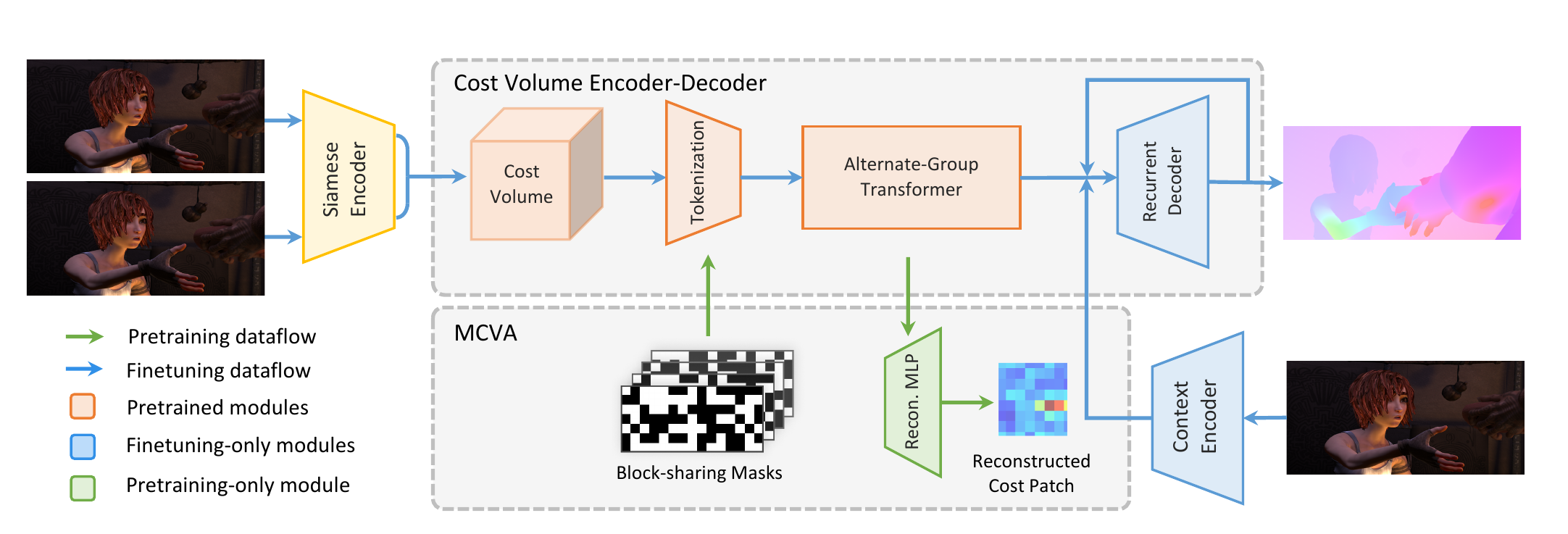}
    \end{center}
    % \vspace{-0.5cm}
    \caption{$\textbf{Overview of FlowFormer with Masked Cost Volume Autoencoding (MCVA).}$ 
    We propose a transformer-based neural network architecture, dubbed as FlowFormer, along with a Masked Cost Volume Autoencoding (MCVA) for optical flow estimation.    
    % The core component of FlowFormer architecture is the transformer-based cost-volume encoder. To unleash the capacity of the transformer encoder, We further propose Masked Cost Volume Autoencoding to pretrain the cost-volume encoder. 
    During pretraining, a portion of cost values are masked and the cost-volume encoder is required to reconstruct masked cost patches.
    In the supervised finetuning stage, FlowFormer recurrently decodes flow from the encoded cost features.
    } 
    \label{fig:figure-overview}
    % \vspace{-0.4cm}
\end{figure*}

Optical flow targets at estimating per-pixel correspondences between a source image and a target image, in the form of a 2D displacement field.
In many downstream video tasks, such as action recognition~\cite{sun2018optical,piergiovanni2019representation,zhao2020improved}, video inpainting~\cite{kim2019deep,xu2019deep,gao2020flow}, video super-resolution~\cite{lai2017deep,8967249,chan2021basicvsr,sajjadi2018frame}, and frame interpolation~\cite{xu2019quadratic,bao2019depth,liu2020video,niklaus2020softmax,huang2020rife}, optical flow serves as a fundamental component providing dense correspondences as valuable clues for prediction.

A general assumption adopted in optical flow estimation is that the appearance of corresponding locations in the two images induced from optical flows remains unchanged. Traditionally, optical flow is modeled as an optimization problem that maximizes visual similarities between cross-image corresponding locations with regularization terms. With the rapid development of deep learning and emerging training data, this field has been significantly advanced by deep convolutional neural network-based methods. The recent methods compute costs (i.e. visual similarities) between feature pairs, upon which flows are regressed. Most successful architecture designs in optical flow are achieved via better designs of cost encoding and decoding. PWC-Net \cite{sun2018pwc} and RAFT \cite{teed2020raft} are two recent representative deep learning-based methods. PWC-Net \cite{sun2018pwc} builds hierarchical local cost volumes with warped features and progressively estimates flows from such local costs. RAFT \cite{teed2020raft} forms an $H\times W\times H\times W$ 4D cost volume that measures similarities between all pairs of pixels of the pair of $H \times W$ images and iteratively retrieves local costs within local windows for regressing flow residuals.

Recently, transformers have attracted much attention for their ability to model long-range relations.
Network pretraining further unleashes the capacity of transformers with unlabeled data, which is fascinating for optical flow because the ground truth collection for optical flow is expensive.
Perceiver IO \cite{jaegle2021perceiver} is the pioneering work that learns optical flow regression with a transformer-based architecture. 
However, it directly operates on pixels of image pairs and ignores the well-established domain knowledge of encoding visual similarities to cost volumes or maps for flow estimation. It thus requires a large number of parameters and training examples to capture the desired input-output mapping.
Besides, how to pretrain Perceiver IO still remains a challenge.
We therefore raise a question: can we design a transformer architecture along with a pertaining technique for optical flow?
% that enjoys both advantages of transformers and cost volume and can be effectively pretrained?
We propose FlowFormer as an answer to this question. 
% enjoy both advantages of transformers and the cost volume from the previous milestone architectures? Such a question calls for designing novel transformer architectures for optical flow estimation that can effectively aggregate information from the cost volume. In this paper, we introduce the novel optical Flow TransFormer~(FlowFormer) to address this challenging problem.
FlowFormer contains two main novel techniques: 1) an optical Flow Transformer~(FlowFormer) architecture, and 2) a Masked Cost Volume Autoencoding (MCVA) technique to pretrain the proposed FlowFormer.

FlowFormer directly adopts an ImageNet-pretrained Twins~\cite{jiang2021learning} transformer to encode image features, and then uses an asymmetric encoder-decoder architecture for cost-volume encoding and decoding (Fig.~\ref{fig:figure-overview}). After building a 4D cost volume from image features, FlowFormer consists of two main components: 1) a cost-volume encoder that embeds the 4D cost volume into a latent cost space and fully encodes the cost information in such a space, and 2) a lightweight recurrent cost decoder that estimates ﬂows from the encoded latent cost features. Compared with previous works, the main characteristic of our FlowFormer is to adapt the transformer architectures to effectively process cost volumes, which are compact yet rich representations widely explored in optical flow estimation communities, for estimating accurate optical flows.
% Due to the asymmetric architecture design, we can pretrain the cost-volume encoder

A naive strategy to transform the 4D cost volume with transformers is directly tokenizing the 4D cost volume and applying transformers. However, such a strategy needs to use thousands of tokens, which is computationally unbearable. To tackle this challenge, we propose two key designs in our cost encoder.
We propose a two-step tokenization: 1) converting each of the 2D cost maps, which records visual similarities between one source pixel and all target pixels, from the 4D cost volume into patches as commonly done in transformer networks, and 2) further projecting cost-map patches of each cost map into $K$ latent cost tokens. In this way, 
%each 2D cost map is summarized as $K$ latent cost tokens and 
the $H\times W \times H \times W$ 4D cost volume can be transformed into $H\times W\times K$ tokens.
Secondly, instead of performing self-attention among all tokens, we alternatively conduct attention over tokens within the same cost map and tokens across different cost maps. In other words, an interweaving stack of aggregations of latent cost tokens belonging to the same source pixel and those across different source pixels. 
Combining these two designs, FlowFormer encodes the cost volume into compact and globally aware latent cost tokens, dubbed as the {\it cost memory}.

Classical transformer architectures~\cite{carion2020end} decodes information from the encoded memory via stacked cross-attention layers. In contrast to them, inspired by RAFT, our cost decoder adopts only a recurrent attention layer that formulates the cost decoding as a recurrent query process with dynamic positional cost queries: based on current estimated flows, we query the cost memory for regressing the flow residuals.
In each iteration, we compute the corresponding positions in the target image for all source pixels according to current flows and then dynamically update positional cost queries with such positions.
% cost queries with project such positions into high-dimensional queries~(positional queries) through positional encoding. 
Then, we fetch cost features from the cost memory via cross-attention and use a shared gated recurrent unit~(GRU) head for residual flow regression.

 % The core of its success lies on two aspects: the ImageNet-pretrained transformer-based image encoder and the transformer-based cost-volume encoder. Notably, adopting an ImageNet-pretrained visual backbone leads to considerable performance gain over the train-from-scratch counterpart, indicating that random weight initialization hinders the learning of correspondence estimation. This naturally begs the question: can we also pretrain the transformer-based cost-volume encoder and thus further unleash its power to achieve more accurate optical flow?

We also propose masked cost-volume autoencoding (MCVA), a self-supervised pretraining scheme to enhance the cost-volume encoding on top of the FlowFormer framework. We are inspired by the recent success of masked autoencoding, such as BERT~\cite{devlin2018bert} in NLP and MAE~\cite{he2022masked} in computer vision. 
The key idea of masked autoencoding is masking a portion of input data, and requiring networks to learn high-level representation for masked contents reconstruction.
Following this paradigm, we pretrain the cost-volume encoder by reconstructing the masked cost volume with another reconstruction head, as shown in Fig.~\ref{fig:figure-overview}.
% we mask a portion the cost volume and encourage the cost-volume encoder to learn high-level representation by reconstructing the masked contents.
However, it is non-trivial to adapt the masked autoencoding strategy to learn a better cost-volume encoder for optical flow estimation, because of the two following reasons. Firstly, the cost volume might contain redundancy and the cost maps (cost values between a source-image pixel to all target-image pixels) of neighboring source-image pixels are highly correlated. Randomly masking cost values~\cite{he2022masked} can lead to information leakage and makes the model biased towards aggregating local information. Secondly, existing masked autoencoding methods target at reconstructing masked content randomly selected from fixed locations. This suffices to pretrain general-purpose single-image encoder in other fields. However, the cost-volume encoder of FlowFormer is deeply coupled with the follow-up recurrent decoder, which demands cost information of long range at flexible locations. 

To tackle the aforementioned issues, we introduce two task-specific designs. Firstly, instead of randomly masking the cost volume, we partition source pixels into blocks and let source pixels within the same block share a common mask on their cost maps. This strategy, termed block-sharing masking, prevents the cost-volume encoder from reconstructing masked cost values by simply copying from neighboring source pixels' cost maps
% \sout{, and encourages long-range information aggregation by considering useful cues from cost maps belonging to far-away source pixels. }
% Such design enfoces the cost-volume encoder to abstract useful cues from cost maps belonging to far-away source pixels, which encourages long-range information aggregation.
Secondly, to mimic the decoding process in finetuning and thus avoid pretraining-finetuning discrepancy, we propose a novel pre-text reconstruction task as shown in Fig.~\ref{fig:figure-overview}: small cost patches (of shape $9\times9$) are randomly cropped from the cost maps to retrieve features from the cost-volume encoder, aiming to reconstruct larger cost map patches (of shape $15\times15$) centered at the same locations. This is in line with the decoding process of FlowFormer in the finetuning stage.
% This pre-text task explicitly encourages the cost-volume encoder to capture 
% long-range information for cost-volume encoding, which is critical for optical flow estimation. 
Besides, we empirically show that the image encoder, upon which the cost volume is built, should be frozen during pretraining to avoid training collapse.

In essence, the proposed masked cost-volume autoencoding (MCVA) has unique designs compared with conventional MAE methods, which encourages the cost-volume encoder 1) to construct high-level holistic representation of the cost volume, more effectively encoding long-range information, 2) to reason about occluded (\textit{i.e.}, masked) information by aggregating faithful unmasked costs, and 3) to decode task-specific feature (\textit{i.e.}, larger cost patches at required locations) to better align the pretraining process with that of the finetuning. 
% These designs contribute to better handling of hard cases, such as noises, large-displacement motion and occlusion, for more accurate flow estimation.

Our contributions can be summarized as sixfold. 1) We propose a novel transformer-based neural network architecture, FlowFormer, for optical flow estimation, which achieves state-of-the-art flow estimation performance. 2) We design a novel cost-volume encoder, effectively aggregating cost information into compact latent cost tokens.
3) We propose a recurrent cost decoder that recurrently decodes cost features with dynamic positional cost queries to iteratively refine the estimated optical flows.
4) We propose the masked cost-volume autoencoding scheme to better pretrain the cost-volume encoder of FlowFormer. 5) We propose task-specific masking strategy and reconstruction pre-text task to mitigate pretraining-finetuning discrepancy, fully taking advantage of the learned representations from pretraining. 6) With the proposed pretraining technique, FlowFormer+MCVA obtains all-sided improvements over FlowFormer, setting new state-of-the-art performance on public benchmarks.

\section{Related Works}
\textbf{Optical Flow.} Traditionally, optical flow was modeled as an optimization problem that maximizes visual similarity between image pairs with regularizations \cite{horn1981determining,black1993framework,bruhn2005lucas,sun2014quantitative}. Major improvements in this era came from better designs of similarity and regularization terms.
The rise of deep neural networks significantly advanced this field. FlowNet \cite{dosovitskiy2015flownet} was the first end-to-end convolutional network for optical flow estimation. Its successive work, FlowNet2.0 \cite{ilg2017flownet}, adopted a stacked architecture with warping operation, performing on par with state-of-the-art (SOTA) methods. 
Then a series of works, represented by SpyNet \cite{ranjan2017optical}, PWC-Net \cite{sun2018pwc,sun2019models}, LiteFlowNet \cite{hui2018liteflownet,hui2020lightweight} and VCN \cite{yang2019volumetric}, employed coarse-to-fine and iterative estimation methodology. These models inherently suffered from missing small fast-motion objects in coarse stage. To remedy this issue, Teed and Deng \cite{teed2020raft} proposed RAFT~\cite{teed2020raft}, which performs optical flow estimation in a coarse-and-fine (i.e. multi-scale search window in each iteration) and recurrent manner.
Based on RAFT architecture, many works \cite{jiang2021learning,xu2021high,jiang2021learning2,zhang2021separable,hofinger2020improving} were proposed to either reduce the computational costs or improve the flow accuracy.
Recently, optical flow was extended to more challenging settings, such as low-light~\cite{zheng2020optical}, foggy~\cite{yan2020optical}, and lighting variations~\cite{huang2021life}.

Among these explorations, visual similarity is computed by the correlation of high dimensional features encoded by a convolutional neural network, and the cost volume that contains visual similarity of pixels pairs acts as a core component supporting optical flow estimation.
However, their cost information utilization lacks effectiveness.
% However, these architectures directly retrieve local costs for flow refinement.
% Zhang et.al.~\cite{} refined the cost volume with aggregation modules introduced by GA-Net~\cite{zhang2019ga}, a stereo matching method, but still lacked efficiency.
We propose FlowFormer that aggregates the cost volume in a latent space with transformers~\cite{vaswani2017attention}.
% one of the major issue that these architectures
% suffering from is the limitation of neural netowrk perceptive field.
% The CNN used to encode features for building cost volume only look at local regions and the they only refine flows from local costs around current flows.
Perceiver IO~\cite{jaegle2021perceiver} pioneered the use of transformers~\cite{vaswani2017attention,dosovitskiy2020image,carion2020end} that is able to establish long-range relationships in optical flow and achieved state-of-the-art performance. 
It ignored the cost volume, showing the strong expressive capacity of transformer architecture at the cost of $\sim80\times$ training examples.
In contrast, we propose to keep cost volume as a compact similarity representation and push search space to the extreme by globally aggregating similarity information via a transformer architecture.
Such global encoding operation is especially beneficial in the hard cases of large displacement and occlusion.
There are also other optical flow methods~\cite{xu2022gmflow,sui2022craft} enhancing image feature encoder with transformers but their performance is not competitive.

\noindent\textbf{Transformers for Computer Vision.} Transformers achieved great success in Natural Language Processing \cite{vaswani2017attention,dai2019transformer,devlin2018bert}, which inspired the development of self-attention for image classification~\cite{liu2021swin,dosovitskiy2020image,chu2021twins}.
Since then, transformer-based architectures has been introduced into many other vision tasks, such as detection \cite{carion2020end}, point cloud processing \cite{guo2021pct,zhao2021point}, image restoration \cite{chen2021pre,liang2021swinir}, video inpainting\cite{zeng2020learning,liu2021fuseformer}, etc, and achieves state-of-the-art in most tasks. The appealing performance is generally attributed to the long-range modeling capacity, which is also a desired property in optical flow estimation.
One of the challenges that vision transformers are faced with is the large number of visual tokens because the computational cost quadratically increases along with the token number.
Twins~\cite{chu2021twins} proposed a spatially separable self-attention (SS Self-Attention) layer that propagates information over tokens arranged in a 2D plane.
We also adopt the SS Self-Attention in the cost-volume encoder to propagate information inter-cost-maps.
Perceiver IO~\cite{jaegle2021perceiver} proposed a general transformer backbone, which although requires a large amount of parameters, achieves state-of-the-art optical flow performance.
Visual correspondence tasks~\cite{sun2021loftr,huang2021vs,cho2021cats,jiang2021cotr} is a main stream in computer vision.
Recently, transformers also lead a trend in such tasks, which is more related to ours. For the sparse matching problem, SuperGlue \cite{sarlin2020superglue} introduced a flexible context aggregation mechanism based on attention. LoFTR \cite{sun2021loftr} utilized transformer to remove feature detector and further improved performance. For semantic correspondence task, CATs \cite{cho2021cats} applied transformer over the cost volume to explore global consensus. However, it directly applied attention on raw cost maps, making it restricted to low-resolution input images.
In contrast, our FlowFormer projects the cost volume into a latent space that is more compact and efficient.
COTR \cite{jiang2021cotr} reformulated dense correspondence problem as functional mapping and took coordinates as input. But its accuracy highly relied on repeated "zoom-in" query operation, which hinders its use for full-image query as required by optical flow estimation. 

\begin{figure*}[t!]
    \centering
    \resizebox{0.95\linewidth}{!}{
\includegraphics[width=\linewidth, trim={5mm 80mm 230mm 0mm}, clip]{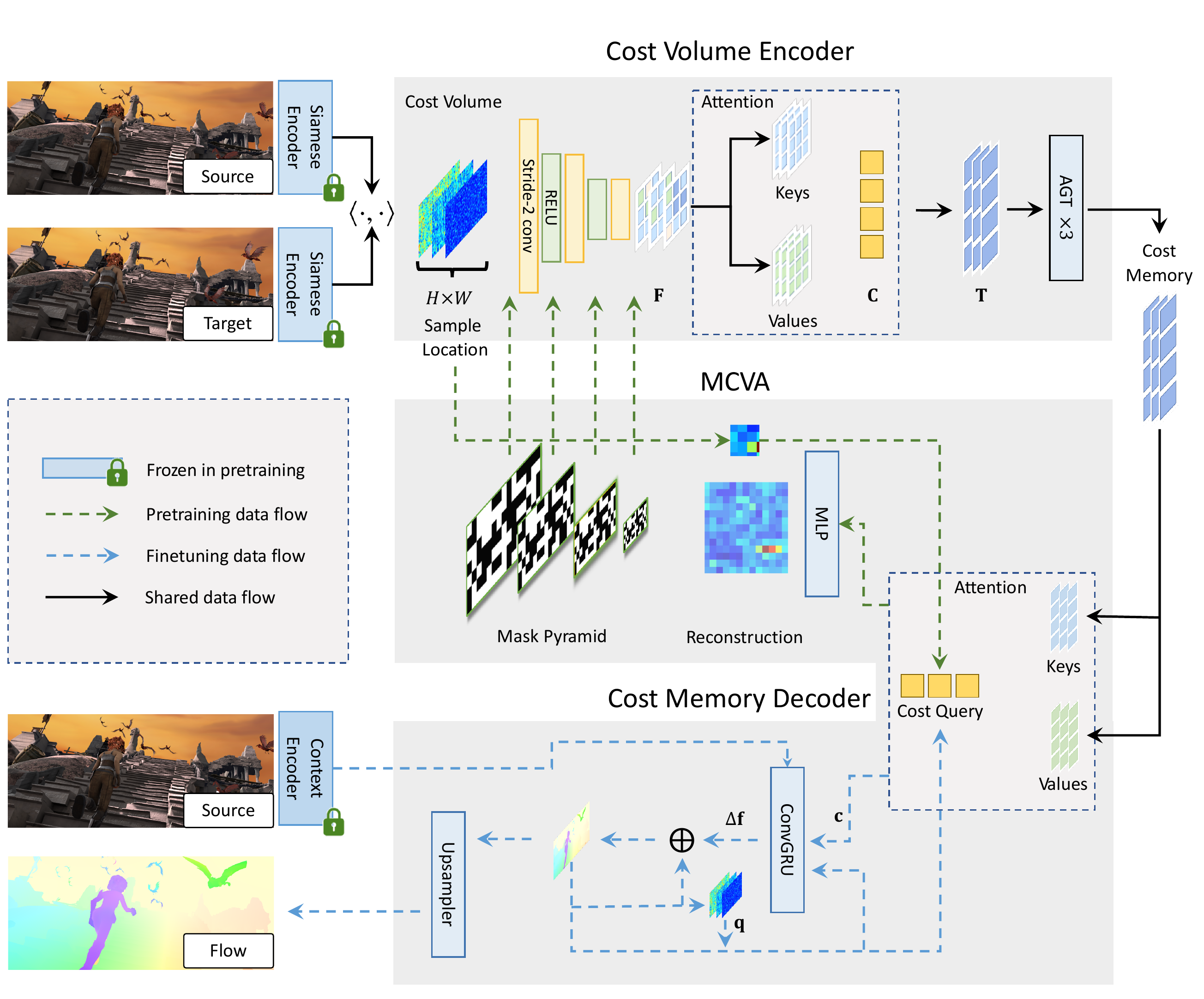}
    % \begin{subfigure}[b]{0.44\linewidth}
        % \includegraphics[width=\linewidth, trim={0mm 30mm 75mm 0}, clip]{figures/architecture.pdf}
        % \caption{\small Attention of projection}
    % \end{subfigure}
    }
    \caption{\textbf{FlowFormer with MCVA}. FlowFormer estimates optical flow in three steps: 1) building a 4D cost volume from image features;
    2) A cost-volume encoder that encodes the cost volume into the cost memory; 3) A recurrent transformer decoder that decodes the cost memory with the source image context features into flows. During MCVA pertaining, FlowFormer freezes the image and context encoders, block-wisely masks the cost volume, and learns to reconstruct larger cost patches from small cost patches to pretrain cost-volume encoder. 
    }
    \label{Fig: arc}
\end{figure*}

\noindent \textbf{Masked Autoencoding (MAE).} As a self-supervised learning technique, MAE, \eg, BERT~\cite{devlin2018bert}, achieved great success in NLP. 
Based on transformers, they mask a portion of the input tokens and require the models to predict the missing content from the reserved tokens.
Pretraining with MAE encourages transformers to build effective long-range feature relationships.
Recently, transformers also stream into the computer vision area, such as image recognition~\cite{liu2021swin,dosovitskiy2020image,chu2021twins}, video inpainting~\cite{zeng2020learning,liu2021fuseformer}, optical flow~\cite{xu2022gmflow}, point cloud recognition~\cite{guo2021pct,zhao2021point}.
By breaking the limitations that convolution can only model local features, transformers present a significant performance gap compared to the previous counterparts.
Pretraining with MAE is also introduced to these modalities, \textit{e.g.}, image~\cite{chen2020generative,he2022masked,xie2022simmim,gao2022convmae}, video~\cite{tong2022videomae}, point cloud~\cite{yu2022point,pang2022masked,hess2023masked,zhang2022point}.
% With transformers came the masked autoencoding scheme.
% A series of recent works, such as MAE~\cite{he2022masked}, SimMIM~\cite{xie2022simmim}, ConvMAE~\cite{gao2022convmae}, PointBERT~\cite{yu2022point}, PointMAE~\cite{pang2022masked}, etc,
These works show that MAE effectively releases the transformer power and do not require extra labeled data.
FlowFormer presents a transformer-based cost-volume encoder and achieves state-of-the-art accuracy.
We further propose the masked cost-volume autoencoding to pretrain the cost-volume encoder on a video dataset, which further unleashes the power of the transformer-based cost-volume encoder.

\section{Method}

Our FlowFormer contains two main parts: an optical Flow Transformer (FlowFormer) and a Masked Cost Volume Autoencoding (MCVA) technique to pretrain FlowFormer. We will elaborate them in this section.

\subsection{FlowFormer Architecture}

The task of optical flow estimation requires to output a per-pixel displacement field 
$\mathbf{f}: \mathbb{R}^{2} \rightarrow \mathbb{R}^2$ that maps every 2D location $\mathbf{x} \in \mathbb{R}^{2}$ of a source image $\mathbf{I}_s$ to its corresponding 2D location $\mathbf{p} = \mathbf{x} + \mathbf{f}(\mathbf{x})$ of a target image $\mathbf{I}_t$. 
To take advantage of the recent vision transformer architectures as well as the 4D cost volumes widely utilized by previous CNN-based optical flow estimation methods,
we propose FlowFormer, a transformer-based architecture that encodes and decodes the 4D cost volume to achieve accurate optical flow estimation.
In Fig.~\ref{Fig: arc}, we show the overview architecture of FlowFormer, which processes the 4D cost volumes from siamese features with two main components: 1) a cost-volume encoder that encodes the 4D cost volume into a latent space to form cost memory, and 2) a cost memory decoder for predicting a per-pixel displacement field based on the encoded cost memory and contextual features.

\subsubsection{Building the 4D Cost Volume}

A backbone vision network is used to extract an $H\times W\times D_f$ feature map from an input $H_I\times W_I\times 3$ RGB image, where typically we set $(H, W)=(H_I/8, W_I/8)$.
%Previous optical flow architectures mostly encode image features with convolutions, which have limited perceptive fields. Recently, transformers have been proved able to capture more complex patterns due to their long-range attention. We propose to extract features with transformer backbones in the hope of capturing more informative patterns. However, directly training transformers as feature encoders is not feasible in optical flow tasks because transformers require a large amount of training data to encode effective inductive bias while training data for optical flow is expensive. [CANNOT SEE WHY IT IS NOT FEASIBLE]
%We use the first two stages of ImageNet \cite{russakovsky2015imagenet} pretrained Twins-Large \cite{chu2021twins} as the feature encoder, which the backbone encodes an RGB image to an $H_I/8\times W_I/8\times 256$ feature map, i.e., $(H, W)=(H_I/8, W_I/8)$.
After extracting the feature maps of the source image and the target image, we construct an $H\times W\times H\times W$ 4D cost volume by computing the dot-product similarities between all pixel pairs between the source and target feature maps.

\subsubsection{Cost-Volume Encoder}

To estimate optical flows, the corresponding positions of source images in the target images need to be identified based on source-target visual similarities encoded in the 4D cost volume.
The built 4D cost volume can be viewed as a series of 2D cost maps of size $H \times W$, each of which measures visual similarities between a single source pixel and all target pixels.
We denote the source pixel $\bf x$'s cost map as ${\bf M_x} \in \mathbb{R}^{H \times W}$.
Finding corresponding positions in such cost maps is generally challenging, as there might exist repeated patterns and non-discriminative regions in the two images. 
The task becomes even more challenging when only considering costs from a local window of the map, as previous CNN-based optical flow estimation methods do.
Even for estimating a single source pixel's accurate displacement, it is beneficial to take its contextual source pixels' cost maps into consideration.

\begin{figure*}[t!]
    \centering
    \resizebox{0.95\linewidth}{!}{
    % \begin{subfigure}[b]{0.44\linewidth}
        \includegraphics[width=\linewidth, trim={10mm 125mm 30mm 0}, clip]{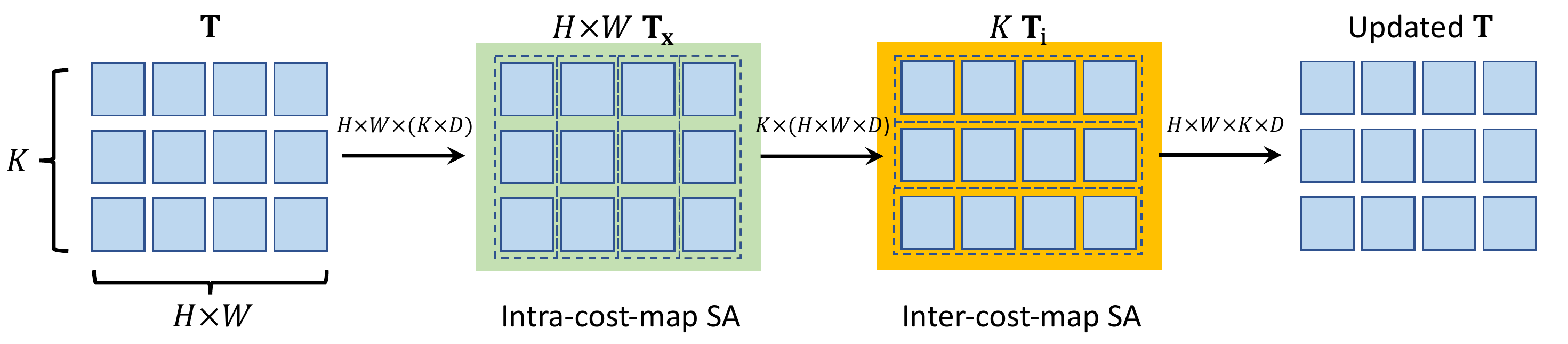}
        % \caption{\small Attention of projection}
    % \end{subfigure}
    }
    \caption{\textbf{Alternate-Group Transformer Layer}. The alternate-group transformer layer~(AGT) alternatively groups tokens in $\mathbf T$ into $H\times W$ groups that contains $K$ tokens ($\mathbf T_\mathbf x$) and $K$ groups that contains $H\times W$ tokens ($\mathbf T_i$), and encode tokens inside groups via self-attention and ss self-attention~\cite{chu2021twins} respectively.
    }
    \label{Fig: agt}
\end{figure*}

To tackle this challenging problem, we propose a transformer-based cost-volume encoder that encodes the whole cost volume into a {\it cost memory}.
Our cost-volume encoder consists of three steps: 1) cost map pathification, 2) cost patch token embedding, and 3) cost memory encoding.
We elaborate the details of the three steps as follows.

\vspace{2pt}
\noindent \textbf{Cost map patchification.} 
%A cost map is of $H_f\times W_f$ whose feature length is of $1$. Inspired by DETR that extract high dimensional features from three channel RGB images with convolutions for the following transformer encoding, 
Following existing vision transformers, we patchify the cost map $\mathbf{M_x} \in \mathbb{R}^{H \times W}$ of each source pixel $\bf x$ with strided convolutions to obtain a sequence of cost patch embeddings.
%feature map with smaller spatial size and higher feature length.
% We patchify cost maps with patch size of 8.
Specifically, given an $H \times W$ cost map, we first pad zeros at its right and bottom sides to make its width and height multiples of 8. 
%A convolution layer with stride 2, padding 2, and kernel size 6 can halve the size of the feature map.
The padded cost map is then transformed by a stack of three stride-2 convolutions followed by ReLU into a feature map $\mathbf{F}_{\bf x} \in \mathbb{R}^{\lceil{H/8}\rceil \times \lceil{W/8}\rceil \times D_p}$.
%, where $H_p= \lceil{H_f/8}\rceil, W_p= \lceil{W_f/8}\rceil$ and 
Each feature in the feature map stands for an $8\times 8$ patch in the input cost map.
The three convolutions have output channels of $D_p/4$, $D_p/2$, $D_p$, respectively.

%design a block that contains three such convolution layer interweaved with two ReLU activation layers. The number of channels of the three convolution layers are $D_p/4$, $D_p/2$, and $D_p$. By applying such block to the padded cost map, we can maintain a $H_p\times H_p\times D_p$ feature map ($H_p= \lceil{H_f/8}\rceil, W_p= \lceil{W_f/8}\rceil$) where each feature stands for a patch of size $8\times 8$ in the original cost map.

\vspace{2pt}
\noindent \textbf{Patch Feature Tokenization via Latent Summarization.}
Although the patchification results in a sequence of cost patch feature vectors for each source pixel, the number of such patch features is still large and hinders the efficiency of information propagation among different source pixels. Actually, a cost map is highly redundant because only a few high costs are most informative. To obtain more compact cost features, we further summarize the patch features $\mathbf F_{\bf x}$ of each source pixel $\mathbf x$ via $K$ latent codewords $\mathbf{C} \in \mathbb{R}^{K\times D}$. Specifically, the latent codewords query each source pixel's cost-patch features to further summarize each cost map into $K$ latent vectors of $D$ dimensions via the dot-product attention mechanism. The latent codewords ${\bf C} \in \mathbb{R}^{K \times D}$
are randomly initialized, updated via back-propagation, and shared across all source pixels. The latent representations ${\bf T_x}$ for summarizing ${\bf F_x}$ are obtained as
\begin{equation}
\begin{aligned}
    % & \mathbf Q_{latent} = \text{MLP}(\mathbf C), \\
    & {\bf K_x} = \text{Conv}_{1\times 1}\left(\text{Concat}({\bf F_x}, {\rm PE})\right), \\
    & {\bf V_x} = \text{Conv}_{1\times 1}\left(\text{Concat}({\bf F_x}, {\rm PE})\right), \\
    & {\bf T_x} = \text{Attention}({\bf C}, {\bf K_x}, {\bf V_x}).
\end{aligned}
\label{Eq: latent projection}
\end{equation}
Before projecting the cost-patch features $\bf F_x$ to obtain keys $\bf K_x$ and values $\bf V_x$, the patch features are concatenated with a sequence of positional embeddings $\text{PE} \in \mathbb{R}^{\lceil{H/8}\rceil\times \lceil{W/8}\rceil\times D_p}$. 
Given a 2D position $\mathbf p$, we encode it into a positional embedding of length $D_p$ following COTR \cite{jiang2021cotr}.
% via Eq.~\ref{Eq: pe}.
%After concatenating the features and the position encodings, we generate keys $\mathbf K_{patch}\in \mathbb{R}^{(H_p\times W_p)\times D}$ and values $\mathbf V_{patch}\in \mathbb{R}^{(H_p\times W_p)\times D}$ with a $1\times1$ convolution.
% \begin{equation}
% \begin{aligned}
%     % & \mathbf Q_{latent} = \text{MLP}(\mathbf C), \\
%     & \mathcal{P}(\mathbf p) = [sin(\pi\mathbf p), cos(\pi \mathbf p), ..., sin(\frac{D}{4}\pi\mathbf p), cos(\frac{D}{4}\pi\mathbf p)]
% \end{aligned}
% \label{Eq: pe}
% \end{equation}
%The PE computes positional embeddings for positions ranging from $(0, 0)$ to $(\lceil{H/8}\rceil, \lceil{W/8}\rceil)$.
Finally, the cost map of the source pixel $\bf x$ can be summarized into $K$ latent representations ${\bf T_x} \in \mathbb{R}^{K\times D}$ by conducting multi-head dot-product attention with the queries, keys, and values. Generally, $K \times D \ll H \times W$ and the latent summarizations $\bf T_x$ therefore provides more compact representations than each $H \times W$ cost map for each source pixel $\bf x$.
%Through the two-step tokenization, the $N$ latent cost codes in $\mathbf C$ can seek for information that they are interested in over the whole cost map and generate $N$ latent cost tokens.
% We visualize the attention of two examples in Fig.~\ref{}.
% For simple cases that the cost map is clean and the highest heat locates at the corresponding position, the $K$ latent cost tokens focus on the patches around the corresponding patch.
% For challenging cases that the cost map is noisy and misleading, the $K$ latent cost tokens absorb information from most potential patches.
For all source pixels in the image, there are a total of $(H\times W)$ 2D cost maps. Their summarized representations can consequently be converted into a latent 4D cost volume ${\bf T} \in \mathbb{R}^{H \times W \times K \times D}$.
%We project all patchified cost maps with the shared $\mathbf{C}$, which produces $\mathbf T \in \mathbb{R}^{H_f\times W_f\times N\times D}$ containing $H_f\times W_f\times N$ latent cost tokens of length $D$.

% $H_f\times W_f$ groups of latent cost tokens and each group contains $N$ tokens.
% and each group contains $K$ .

\vspace{2pt}
\noindent \textbf{Attention in the latent cost space.} The aforementioned two stages transform the original 4D cost volume into a latent and compact 4D cost volume $\bf T$. 
However, it is still too expensive to directly apply self-attention over all the vectors in the 4D volume because the computational cost quadratically increases with the number of tokens. 
As shown in Fig.~\ref{Fig: agt}, we propose an alternate-group transformer layer (AGT) that groups the tokens in two mutually orthogonal manners and apply self-attentions in the two groups alternatively, which reduces the computational cost of self-attention  while still being able to propagate information between all tokens.

The first grouping is conducted for each source pixel, i.e., each ${\bf T_x} \in \mathbb{R}^{K \times D}$ forms a group and the self-attention is conducted within each group.
\begin{equation}
\begin{aligned}
    {\bf T_x} = {\rm FFN}({\rm Self\text{-}Attention}({\bf T_x}(1), \dots, {\bf T_x}(K)) \\ ~~~\text{ for all } {\bf x} \text{ in } {\bf I}_s,
    % & \mathbf K_{patch} = Conv_{1\times 1}(Concat(\mathbf F_p(\mathbf x), PE)), \\
\end{aligned}
\label{Eq: intra-cost}
\end{equation}
where ${\bf T_x}(i)$ denotes the $i$-th latent representation for encoding the source pixel $\bf x$'s cost map. After the self-attention is conducted between all $K$ latent tokens for each source pixel $\bf x$, updated $\bf T_x$ are further transformed by a feed-forward network (FFN) and then re-organized back to form the updated 4D cost volume $\bf T$. Both the self-attention and FFN sub-layers adopt the common designs of residual connection and layer normalization of transformers.
%We use a self-attention layer with a FFN layer shared by all groups.
This self-attention operation propagates the information within each cost map and we name it as intra-cost-map self-attention.

The second way groups all the latent cost tokens ${\bf T} \in \mathbb{R}^{H \times W \times K \times D}$
into $K$ groups according to the $K$ different latent representations. 
Each group would therefore have $(H \times W)$ tokens of dimension $D$ for information propagation in the spatial domain via
the spatially separable self-attention (SS-SelfAttention) proposed in Twins~\cite{chu2021twins},
\begin{equation}
\begin{aligned}
    {\bf T}_i = {\rm FFN}({\rm SS}\text{-}{\rm SelfAttention}({\bf T}_i)) ~~~\text{ for } i = 1,2,\dots, K,
\end{aligned}
\label{Eq: inter-cost}
\end{equation}
where we slightly abuse the notation and denote ${\bf T}_i \in \mathbb{R}^{(H \times W) \times D}$ as the $i$-th group. The updated ${\bf T}_i$'s are then re-organized back to obtain the updated 4D latent cost volume $\bf T$. Moreover, visually similar source pixels should have coherent flows, which has been validated by previous methods~\cite{jiang2021learning,cho2021cats}.
Thus, we integrate appearance affinities between different source pixels into SS-SelfAttention via concatenating the source image's context features $\mathbf{t}$ with the cost tokens when generating queries and keys.
We call this layer inter-cost-map self-attention layer as it is responsible to propagate information of cost volume across different source pixels.

The above self-attention operations' parameters are shared across different groups and they are sequentially operated to form the proposed alternate-group attention layer.
%An AGT layer consisting of an intra-cost-map self-attention layer and an inter-cost-map self-attention layer.
By stacking the alternate-group transformer layer multiple times, the latent cost tokens can effectively exchange information across source pixels and across latent representations to better encode the 4D cost volume.
%, which helps each group of $K$ latent tokens that originally encode corresponding cost maps to absorbs all visual similarity information stored in the whole cost volume.
In this way, our cost-volume encoder transforms the $H\times W\times H\times W$ 4D cost volume to $H\times W\times K$ latent tokens of length $D$.
We call the final $H\times W\times K$ tokens as the \textit{cost memory}, which is to be decoded for optical flow estimation.

\subsubsection{Cost Memory Decoder for Flow Estimation}

Given the cost memory encoded by the cost-volume encoder, we propose a cost memory decoder to predict optical flows. 
Since the original resolution of the input image is $H_I \times W_I$,
we estimate optical flow at the $H\times W$ resolution and then upsample the predicted flows to the original resolution with a learnable convex upsampler \cite{teed2020raft}.
However, in contrast to previous vision transformers that seek abstract semantic features, optical flow estimation requires recovering dense correspondences from the cost memory.
Inspired by RAFT \cite{teed2020raft}, we propose to use cost queries to retrieve cost features from the cost memory and
iteratively refine flow predictions with a recurrent attention decoder layer.

\vspace{2pt}
\noindent \textbf{Cost memory aggregation.}
For predicting the flows of the $H\times W$ source pixels, we generate a sequence of $(H \times W)$ cost queries, each of which is responsible for estimating the flow of a single source pixel via co-attention on the cost memory. To generate the cost query $\bf Q_x$ for a source pixel $\bf x$, we first compute its corresponding location in the target image given its current estimated flow ${\bf f}({\bf x})$ as ${\bf p_x} = {\bf x} + {\bf f}({\bf x})$. 
We then retrieve a local $9\times 9$ cost-map patch $\mathbf{q_x}=\mathrm{Crop}_{9\times 9}(\mathbf{M_x}, \mathbf{p_x})$ by cropping costs inside the $9\times 9$ local window centered at $\mathbf{p_x}$ on the cost map $\mathbf{M_x}$.
The cost query $\bf Q_x$ is then formulated based on the features $\text{FFN}(\mathbf{q_x})$ that encoded from the local costs $\mathbf{q_x}$ and $\mathbf p_{\mathbf x}$'s positional embedding $\mathrm{PE}({\mathbf p_{\mathbf x}})$, which can aggregate information from source pixel $\bf x$'s cost memory $\bf T_x$ via cross-attention,
\begin{equation}
\begin{aligned}
    & {\bf Q_x} = \text{FFN}\left(\text{FFN}(\mathbf{q_x})  + \mathrm{PE}(\mathbf {p_x}) \right), \\
    & {\bf K_x} = \text{FFN}\left({\bf T_x}\right),~~{\bf V_x} = \text{FFN}\left({\bf T_x}\right), \\
    & {\bf c_x} = \mathrm{Attention}({\bf Q_x}, {\bf K_x}, {\bf V_x}).
\end{aligned}
\label{Eq: decoder attention}
\end{equation}
The cross-attention summarizes information from the cost memory for each source pixel to predict its flow. 
As $\bf Q_x$ is dynamically updated in terms of the fed position at each iteration, we call it as dynamic positional cost query.
We note that keys and values can be generated at the beginning and re-used in subsequent iterations, which saves computation as a benefit of our recurrent decoder.

%Given current flow estimation $\mathbf{f}(\mathbf{x})$ of a source pixel at location $\mathbf{x}$, we compute its corresponding location at the target image $\mathbf{p}={\bf x} + {\bf f}({\bf x})$ and cost query feature $\mathbf{q}_{\mathbf x}$ with local costs inside the $9\times 9$ window centered at $\mathbf p$.
%By encoding $\mathbf p$ into a high dimensional positional embedding $\mathcal{P}(\mathbf p)$, we add the positional embedding to the cost query feature and then compute a cost query $\mathbf{Q}_{\mathbf{x}}$ with a FFN.
%We collect the $K$ tokens that correspond to $\mathbf x$ in the cost memory, ${\mathbf T_\mathbf x}$ and compute keys ${\mathbf K_\mathbf x}$ and values ${\mathbf V_\mathbf x}$ from them with two FFNs.
%Finally, we extract cost information ${\mathbf c_\mathbf x}$ that stored in ${\mathbf T_\mathbf x}$ through attention.

% \begin{figure}[t!]
%     \centering
%     \resizebox{1.0\linewidth}{!}{
%     % \begin{subfigure}[b]{0.44\linewidth}
%         \includegraphics[width=\linewidth, trim={35mm 65mm 60mm 40mm}, clip]{Figures/decoder.pdf}
%         % \caption{\small Attention of projection}
%     % \end{subfigure}
%     }
%     \caption{Decoder.
%     }
%     \label{Fig: arc}
% \end{figure}

\vspace{3pt}
\noindent \textbf{Recurrent flow prediction.} 
Our cost decoder iteratively regresses flow residuals $\Delta \mathbf f(\mathbf x)$ to refine the flow of each source pixel $\mathbf x$ as $\mathbf f(\mathbf x) \leftarrow \mathbf f(\mathbf x) + \Delta \mathbf f(\mathbf x)$.
We adopt a $\text{ConvGRU}$ module and follow the similar design to that in GMA-RAFT~\cite{jiang2021learning} for flow refinement. 
However, the key difference of our recurrent module is the use of cost queries to adaptively aggregate information from the cost memory for more accurate flow estimation.
Specifically, at each iteration, the ConvGRU unit takes as input the concatenation of retrieved cost features and cost-map patch $\mathrm{Concat}({\bf c_x, q_x})$, the source-image context feature $\bf t_x$ from the context network, and the current estimated flow $\mathbf f$, and outputs the predicted flow residuals as follows,
\begin{equation}
\begin{aligned}
    & \Delta \mathbf f(\mathbf x) = \mathrm{ConvGRU}(\mathrm{Concat}({\bf c_x}, {\bf q_x}), {\bf t_x}, \mathbf f(\mathbf x)).
\end{aligned}
\label{Eq: decoder gru}
\end{equation}
The flows generated at each iteration are unsampled to the size of the source image via a convex upsampler following \cite{teed2020raft} and supervised by ground-truth flows at all recurrent iterations with increasing weights.

% \begin{figure*}[t!]
%     \centering
%     \resizebox{1.0\linewidth}{!}{
%         \includegraphics[width=\linewidth, trim={5mm 80mm 230mm 0mm}, clip]{figures/arc.pdf}
%     }
% \caption{$\textbf{Overview of Masked Cost Volume Autoencoding.}$ During pertaining, MCVA-FlowFormer freezes the image and context encoders, block-wisely masks the cost volume, and learns to reconstruct larger cost patches from small cost patches to pretrain cost-volume encoder. 
%     In fine-tuning, MCVA-FlowFormer uses the full cost volume, removes the reconstruction decoder, and adds the cost memory decoder to learn optical flow, which naturally falls back to the FlowFormer architecture but inherits the pretrained parameters in the cost-volume encoder.
%     }
%     \label{Fig: arc}
%     \vspace{-0.4cm}
% \end{figure*}

% In this section, we introduce the proposed masked cost-volume autoencoding (MCVA) scheme. We first illustrate the masking strategy that prevents information leakage in pretraining. Then we explain how to modify cost-volume embedding modules to make them compatible with masks. Finally, we present proposed pre-text reconstruction task which minimizes pretraining-fituning discrepancy.

\subsection{Masked Cost Volume Autoencoding}

To better unleash the potential of FlowFormer architecture, as presented in Fig.~\ref{Fig: arc}, we propose a masked cost-volume autoencoding (MCVA) scheme to pretrain the cost-volume encoder of FlowFormer framework.
The key of general masked autoencoding methods is to mask a portion of data and encourage the network to reconstruct the masked tokens from visible ones. 
% \sout{The MCVA adopts this paradigm to cost-volume encoder with three key designss: a proper masking strategy on the cost volume, modifying FlowFormer architecture to make it compatible with masks, and a novel pre-text reconstruction task supervising the pre-training process. }
Due to the redundant nature of the cost volume and the original FlowFormer architecture being incompatible with masks, naively adopting this paradigm to pretrain the cost-volume encoder leads to inferior performance. Our proposed MCVA tackles the challenge and conducts masked autoencoding with three key components: a proper masking strategy on the cost volume, modifying FlowFormer architecture to accommodate masks, and a novel pre-text reconstruction task supervising the pretraining process. 

% The critical two components in the MCVA are the cost volume masking strategy, and how to modify the architecture of FlowFormer to make it compatible with masks.
In this subsection, we 
% elaborate the proposed three key designs.
%elaborate our adaptations on modules of the cost-volume encoder to perform MCVA scheme.
first introduce the masking strategy, dubbed as block-sharing masking, and then show the masked cost-volume tokenization that makes the cost-volume encoder compatible with masks. Coupling these two designs prevents the masked autoencoding from being hindered by information leakage in pretraining. Finally, we present the pre-text cost reconstruction task, mimicing the decoding process in finetuning to pretrain the cost-volume encoder.

\begin{figure}[!t]
    \begin{center}
    \includegraphics[width=1\linewidth]{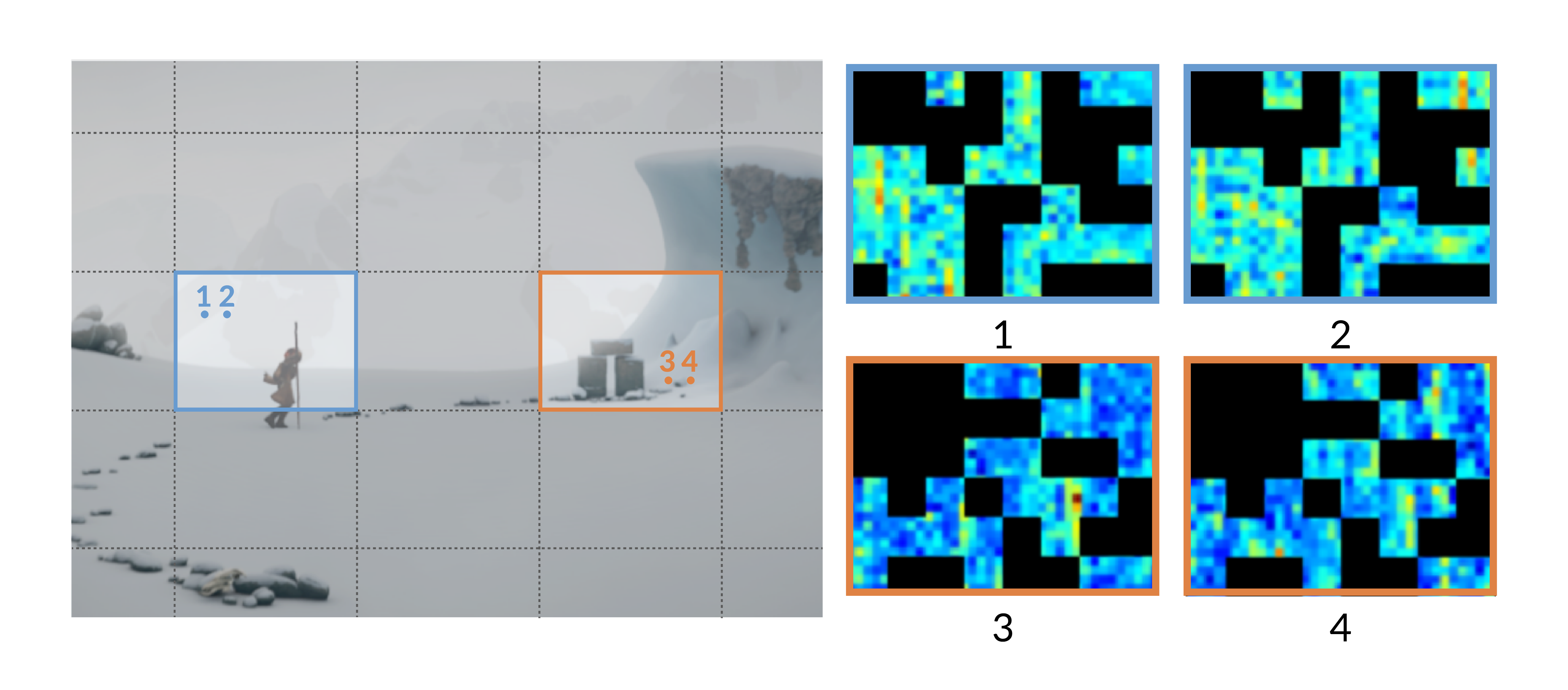}
    \end{center}
    \caption{$\textbf{Block-sharing masking.}$ Source pixels within the same block share a common mask.}
    \label{fig: block-sharing masking}
\end{figure}

\subsubsection{Block-sharing Cost Volume Masking}

% The cost tokenizer and the cost-volume encoder AGT encode the cost memory from the 4D cost volume.

%In pretraining, we build a jarless cost volume from an image pair via \textbf{frozen} ImageNet-pretrained Twins-SVT.
%To allow masked cost-volume autoencoding, 
A properly designed masking scheme is required to conduct autoencoding of the masked cost volume. 
% \xy{\sout{We {\bf freeze} the ImageNet-pretrained Twins-SVT backbone and build the cost volume from the pair of input images. INTUITIVE ADVANTAGES OF FREEZING IMAGENET WEIGHTS??? CREATE LEARNING TARGETS?}}
% \xy{\sout {To pre-train the parameters of the cost tokenizer and the alternate-transformer (AGT) layers,
% we randomly mask the cost volume with a high ratio and reconstruct the masked region with a lightweight decoder.}} 
For each source pixel $\mathbf x$, we need to create a binary mask ${\bf B_x} \in \{0,1\}^{H \times W}$ to its cost map ${\bf A_x} \in \mathbb{R}^{H \times W}$, where 0 indicates masking (\textit{i.e.}, removing) cost values from the masked locations.
Naturally, neighboring source pixels' cost maps are highly correlated. Randomly masking neighboring pixels' cost maps might cause information leakage, \textit{i.e.}, masked cost values might be easily reconstructed by copying the cost values from neighboring source pixels' cost maps.

To prevent such an over-simplified learning process, we propose a block-sharing masking strategy (Fig.~\ref{fig: block-sharing masking}). We partition source pixels into non-overlapping blocks in each iteration. All source pixels belonging to the same block share a common mask for masked region reconstruction. In this way, neighboring source pixels are unlikely to copy each other's cost maps to over-simplify the autoencoding process. Besides, the size of block is designed to be large (height and width of blocks are of $32\sim120$ pixels) and randomly changes in each iteration, and thus encouraging the cost-volume encoder to aggregate information from long-range context and to filter noises of cost values. The details of the mask generation algorithm are provided in supplementary.
% \begin{algorithm}
% \caption{Pseudocode for block-sharing mask generation.}
% \label{alg:mask_generation}
% \begin{algorithmic}
% \State {\#PyTorch-style; follow MAE fast implementation}
% \State {//H, W are heigh and width of feature map}
% \State {blockHeight = randInt(4,15)}
% \State {blockWidth = randInt(4,15)}
% \State {//First generate mask at 1/8 resolution}
% \State {noiseValue = rand(H, W, H//8, W//8, 1)}
% \State {noiseValue.repeat(1, 1, 1, 1, blockHeight$\times$blockWidth)}

% \Require $n \geq 0$
% \Ensure $y = x^n$
% \State $y \gets 1$
% \State $X \gets x$
% \State $N \gets n$
% \While{$N \neq 0$}
% \If{$N$ is even}
%     \State $X \gets X \times X$
%     \State $N \gets \frac{N}{2}$  \Comment{This is a comment}
% \ElsIf{$N$ is odd}
%     \State $y \gets y \times X$
%     \State $N \gets N - 1$
% \EndIf
% \EndWhile
% \end{algorithmic}
% \end{algorithm}
Specifically, for each source pixel's cost map, we first generate the mask map ${\mathbf B}_{\mathbf x}^3\in \{0,1\}^{\frac{H}{8} \times \frac{W}{8}}$ at $\frac{1}{8}$ resolution, and then up-sample it $2\times$ for three times to obtain a pyramid of mask maps ${\mathbf B}_{\mathbf x}^i\in \{0,1\}^{\frac{H}{2^i} \times \frac{W}{2^i}}$, where $i \in \{0,1,2\}$, which are used for the down-sampling encoding process and will be discussed later in Sec.~\ref{sec:tokennization}.

Another key design is that, in pretraining, we {\bf freeze} the ImageNet-pretrained Twins-SVT backbone to build the cost volume from the pair of input images. Freezing the image encoder ensures the reconstruction targets (\textit{i.e.}, raw cost values) to maintain static and avoids training collapse.

\subsubsection{Masked Cost-volume Tokenization}
\label{sec:tokennization}

Given the above generated mask for each source pixel ${\bf x}$'s cost map,
FlowFormer adopts a two-step cost-volume tokenization before the cost encoder.
To prevent the masked costs from leaking into subsequent cost aggregation layers, 
the intermediate embeddings of the cost map need to be properly masked in the cost-volume tokenization process.
We propose the masked cost-volume tokenization, which prevents mixing up masked and visible features.
Firstly, FlowFormer patchifies the raw cost map $\mathbf A_{\mathbf x}\in \mathbb R^{H\times W}$ of each source pixel ${\mathbf x}$ (which is obtained by computing dot-product similarities between the source pixel ${\mathbf x}$ and all target pixels) via 3 stacked stride-2 convolutions. We denote the feature maps after each of the 3 convolutions as $\mathbf F_{\mathbf x}^i$, which have spatial sizes of $\frac{H}{2^i}\times \frac{W}{2^i}$ for $i \in \{0,1,2\}$.
We propose to replace the vanilla convolutions used in the FlowFormer with masked convolutions~\cite{gao2022convmae,graham2017submanifold,ren2018sbnet}:
\begin{equation}
    {\mathbf F}_{\mathbf x}^{i+1} = {\rm Conv}_{\rm stride2}\left(\mathrm{ReLU}({\mathbf F}_{\mathbf x}^i\odot \mathbf{ B}_\mathbf{x}^i)\right),
\end{equation}
where $\odot$ indicates element-wise multiplication, $i \in \{0,1,2\}$, 
and ${\mathbf F}_{\mathbf x}^0$ is the raw cost map ${\bf A_x}$. 
The masked convolutions with the three binary mask maps remove all cost features in the masked regions in pretraining.
Secondly, FlowFormer further projects the patchified cost-map features into the latent space via cross-attention. 
We thus remove the tokens in $\mathbf{F}_{\bf x}^3$ indicated by the mask map ${\mathbf B}_{\mathbf x}^3$ and then only project the remaining tokens into the latent space via the same cross-attention.
During finetuning, the mask maps are removed to utilize all cost features, which converts the masked convolution to the vanilla convolution but the pretrained parameters in the convolution kernels and cross-attention layer are maintained.

The masked cost-volume tokenization completes two tasks.
Firstly, it ensures the subsequent cost-volume encoder only processes visible features in pertaining.
Secondly, the network structure is consistent with the standard tokenization of FlowFormer and can directly be used for finetuning so that the pretrained parameters have the same semantic meanings.
After the masked cost-volume tokenization, the cost aggregation layers (\textit{i.e.}, AGT layers) take visible features as input which also don't need to be modified in finetuning. 
The latent features interact with those of other source pixels in AGT layers and are transformed to the cost memory $\mathbf T_{\mathbf x}$. We explain how to decode the cost memory to estimate flows in following section.

\subsubsection{Reconstruction Target for Cost Memory Decoding}
With the masked cost-volume tokenization, the cost encoder encodes the unmasked cost volume into the cost memory.
The next step is decoding and reconstructing the masked regions from the cost memory.
We now formulate the pre-text reconstruction targets, which supervises the decoding process as well as aformentioned embedding and aggregation layers. 
% We start by revisiting the dynamic positional decoding scheme of FlowFormer, and present our reconstruction targets which are highly consistent with the finetuning tasks. 

\begin{figure}[!t]
    \begin{center}
    \includegraphics[width=0.9\linewidth]{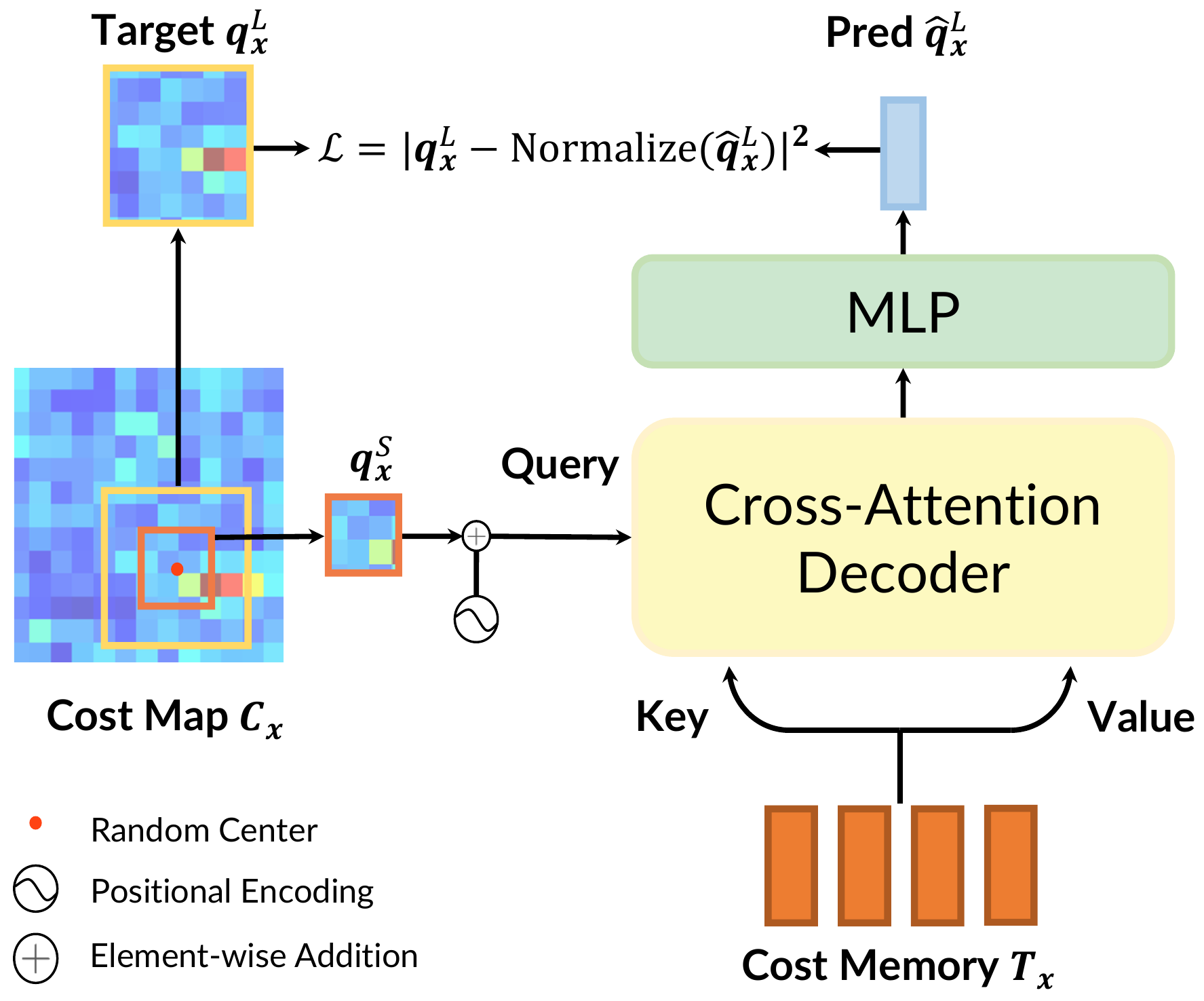}
    \end{center}
    \vspace{-0.5cm}
    \caption{$\textbf{Pre-text reconstruction for cost memory decoding.}$ For each source pixel $\bf x$ and its corresponding cost map $\bf C_x$, a small cost patch ${\bf q}_{\bf x}^S$ is randomly cropped from the cost map to retrieve features from the cost memory $\bf T_x$, aiming to reconstruct larger cost patch ${\bf q}_{\bf x}^L$ centered at the same location.}
    \label{fig:figure1}
    \vspace{-0.4cm}
\end{figure}

\noindent\textbf{Pre-text Reconstruction.}
FlowFormer adopts recurrent flow prediction. In each iteration of the recurrent process, the flow of source pixel ${\mathbf x}$ is decoded from cost memory ${\bf T_x}$, conditioned on current predicted flow, to update the flow prediction. Specifically, current predicted corresponding location in the target image $\bf{p_x}$ is computed as $\bf{p_x=x+f(x)}$, where $\bf{f(x)}$ is current predicted flow. A local cost patch $\bf{q_x}$ is then cropped from the $9 \times 9$ window centered at ${\bf p_x}$ on the raw cost map $\bf{A_x}$. FlowFormer utilizes this local cost patch (with positional encoding) as the query feature to retrieve aggregated cost feature $\bf{c_x}$ via cross-attention operation (Eq.~\ref{Eq: decoder attention}).
Intuitively, ${\bf c_x}$ should contain long-range cost information for better optical flow estimation and it is conditioned on local cost patch $\bf{q_x}$, which indicates the interested location on the cost map. We design a pre-text reconstruction task in line with these two characteristics to pretrain the cost-volume encoder as shown in Fig.~\ref{fig:figure1}: small cost-map patches are randomly cropped from the cost maps to retrieve cost features from the cost memory, targeting at reconstructing larger cost-map patches centered at the same locations.

Specifically, for each source pixel ${\bf x}$, we randomly sample a location ${\bf o_x}$, which is analogous to ${\bf p_x}$ in finetuning. Taking this location as center, we crop a small cost-map patch ${\bf q}_{\bf x}^S=\mathrm{Crop}_{9\times 9}(\mathbf{C_x}, \mathbf{o_x})$ of shape $9 \times 9$. Then we perform the decoding process shown in Eq.~\ref{Eq: decoder attention} to obtain the cost feature ${\bf c_x}$, except that ${\bf p_X}$ and ${\bf q_x}$ are replaced by ${\bf o_x}$ and ${\bf q}_{\bf x}^S$ in pretraining, respectively. To encourage the extracted cost feature ${\bf c_x}$ to carry long-range cost information conditioned on ${\bf o_x}$, we take larger cost-map patch as supervision. Specifically, we crop another larger cost-map patch ${\bf q}_{\bf x}^L=\mathrm{Crop}_{15\times 15}(\mathbf{C_x}, \mathbf{o_x})$ of shape $15 \times 15$ centered at the same location ${\bf o_x}$. We choose a light-weight MLP as prediction head. The MLP takes as input ${\bf c_x}$ and its output ${ \mathbf{\hat q}}_{\bf x}^{L}=\mathrm{MLP}({\bf c_x})$ is supervised by normalized ${\bf q}_{\bf x}^L$. We take mean squared error (MSE) as loss function.
\begin{equation}
    \mathcal{L} = \frac{1}{|\Omega|} \sum_{{\bf x}\in \Omega} \bigg| {\bf q}_{\bf x}^{L} - \mathrm{Normalize}( {\bf{\hat{q}}}_{\bf x}^L)\bigg|^2,
\end{equation} where $\Omega$ is the set of source pixels.

\noindent \textbf{Discussion.}
The key of pretraining is to maintain consistent with finetuning, in terms of both network arthictecture and prediction target.
To this end, we keep the cross-attention decoding layer unchanged and construct inputs with the same semantic meaning (\textit{e.g.}, replacing dynamically predicted ${\bf p_x}$ with randomly sampled ${\bf o_x}$); we supervise the extracted feature ${\bf c_x}$ with long-range cost values to encourage the cost-volume encoder to aggregation global information for better optical flow estimation.
What's more, our scheme only takes an extra light-weight MLP as prediction head, which is unused in finetuning. Compared with previous methods that use a stack of self-attention layers, it is much more computationally efficient.

\section{Experiments}
We denote MCVA pretrained FlowFormer as ``FlowFormer+MCVA''.
We evaluate FlowFormer and FlowFormer+MCVA on the Sintel~\cite{butler2012naturalistic} and KITTI-2015~\cite{geiger2013vision} benchmarks. 
Following previous works, FlowFormer is trained from scratch on FlyingChairs~\cite{dosovitskiy2015flownet} and FlyingThings~\cite{mayer2016large}, and then respectively finetune it on the Sintel and KITTI-2015 benchmarks.
FlowFormer+MCVA is pretrained with Masked Cost-volume Autoencoding on YouTube-VOS~\cite{xu2018youtube} dataset, and then supervised fine-tuning, following the FlowFormer training.
FlowFormer sets new state-of-the-art on the Sintel benchmark and achieves the best generalization performance.
FlowFormer+MCVA further obtains all-sided improvements over FlowFormer, ranking 1st on both benchmarks.

\noindent \textbf{Experimental Setup.}
We adopt the commonly-used average end-point-error (AEPE) as the evaluation metric. It measures the average $l_2$ distance between predictions and ground truth. For the KITTI-2015 dataset, we additionally use the F1-all (\%) metric, which refers to the percentage of pixels whose flow error is larger than 3 pixels or over 5\% of the length of ground truth flows. YouTube-VOS is a large-scale dataset containing video clips from YouTube website. The Sintel dataset is rendered from the same movie in two passes: the clean pass is rendered with easier smooth shading and specular reflections, while the final pass includes motion blur, camera depth-of-field blur and atmospheric effects. The motions in the Sintel dataset are relatively large and complicated. The KITTI-2015 dataset constitutes of real-world driving scenarios with sparse ground truth.

\noindent \textbf{Implementation Details}.
The image feature encoder of our final FlowFormer is chosen as the first two stages of ImageNet-pretrained Twins-SVT~\cite{chu2021twins}, which encodes an image into $D_f=256$-channel feature map of 1/8 image size. The cost-volume encoder patchifies each cost map to a $D_p=64$-channel feature map and further summarizes the feature map to $N=8$ cost tokens of $K=128$ dimensions.
Then, the cost-volume encoder encodes the cost tokens with 3 AGT layers.
Following previous optical flow training procedure~\cite{jiang2021learning}, we train FlowFormer on FlyingChairs~\cite{dosovitskiy2015flownet} for 120k iterations with a batch size of 8, and on FlyingThings~\cite{mayer2016large} for 120k iterations with a batch size of 6~(denoted as `C+T').
Then, we finetune FlowFormer on the data combined from FlyingThings, Sintel, KITTI-2015, and HD1K~\cite{kondermann2016hci}~(denoted as `C+T+S+K+H') for 120k iterations with a batch size of 6.
To achieve the best performance on the KITTI benchmark, we also further finetune FlowFormer on the KITTI-2015 for 50k iterations with a batch size of 6.
We use the one-cycle learning rate scheduler. The highest learning rate is set as $2.5\times 10^{-4}$ on FlyingChairs and $1.25\times 10^{-4}$ on the other training sets.
As positional encodings used in transformers are sensitive to image size, we crop the image pairs for flow estimation and tile them to obtain complete flows following Perceiver IO~\cite{jaegle2021perceiver}. 
We use the tile technique for evaluating optical flow on KITTI because the size of images in KITTI is quite different from training image size.
We use fixed Gaussian weights for tile.
% which will be detailed in the supplementary materials.
FlowFormer+MCVA uses the same architecture as FlowFormer. The image feature encoder and context feature encoder are frozen in pretraining. We pretrain FlowFormer+MCVA on YouTube-VOS for 50k iterations with a batch size of 24. The highest learning rate is set as $5\times 10^{-4}$. During finetuning, we follow the same training procedure of FlowFormer.

\noindent \textbf{Gaussian tile.}
Since positional encodings used in transformers are sensitive to image size and the size of an image pair for test $(H_{test}\times W_{test})$ might be different from those of the training images, $(H_{train}\times W_{train})$,
we crop the test image pair according to the training size and estimate flows for patch pairs separately, and then tile the flows to obtain a complete flow map following a similar strategy proposed in Perceiver IO \cite{jaegle2021perceiver}.
Specifically, we crop the image pair into four evenly-spaced tiles, i.e., $H_{train}\times W_{train}$ image tiles starting at $(0, 0)$, $(0, W_{test}-W_{train})$, $(H_{test}-H_{train}, 0)$, and $(H_{test}-H_{train}, W_{test}-W_{train})$, respectively. For each pixel that is covered by several tiles, we compute its output flow $\bf f$ by blending the predicted flows ${\bf f}_i$ with weighted averaging:
\begin{equation}
\begin{aligned}
    {\bf f} = \frac{\sum_i  w_i {\bf f}_i}{\sum_i w_i},
\end{aligned}
\label{Eq: weighted average}
\end{equation}
where $w_i$ is the weight of the $i$-th tile for the pixel.
We compute the $H_{train}\times W_{train}$ weight map according to pixels' normalized distances $d_{i}$ to the tile center:
\begin{equation}
\begin{aligned}
& d_{i}=||(\frac{u_i}{H_{train}}-0.5, \frac{v_i}{W_{train}}-0.5)||_2, \\
& w_{i}=g(d_{i};\mu=0, \sigma=0.05),
\end{aligned}
\label{Eq: gaussian}
\end{equation}
where $(u_i, v_i)$ denote a pixel $i$'s 2D coordinate.
We use a Gaussian-like $g$ as the weighting function to obtain smoothly blended results. We use this weight map for all the tiles.

\begin{table*}[t]
\begin{center}
% \scriptsize
% \tiny
% \small
% \resizebox{1.0\linewidth}{!}{
\begin{tabular}{clccccccc}
\hline
\multicolumn{1}{c}{\multirow{3}{*}{Training Data}} & \multicolumn{1}{c}{\multirow{3}{*}{Method}} & \multicolumn{2}{c}{Sintel (train)}                    & \multicolumn{2}{c}{KITTI-15 (train)}                    & \multicolumn{2}{c}{Sintel (test)}                     & \multicolumn{1}{c}{KITTI-15 (test)} \\
\cmidrule(r{1.0ex}){3-4}\cmidrule(r{1.0ex}){5-6}\cmidrule(r{1.0ex}){7-8} \cmidrule(r{1.0ex}){9-9} 
\multicolumn{1}{c}{}                               & \multicolumn{1}{c}{}                        & \multicolumn{1}{c}{Clean} & \multicolumn{1}{c}{Final} & \multicolumn{1}{c}{F1-epe} & \multicolumn{1}{c}{F1-all} & \multicolumn{1}{c}{Clean} & \multicolumn{1}{c}{Final} & \multicolumn{1}{c}{F1-all} \\ 
\hline
\multirow{3}{*}{A}                         &    Perceiver IO~\cite{jaegle2021perceiver}   & 1.81 & 2.42 & 4.98  & - & - & - & - \\ 
&    PWC-Net~\cite{sun2018pwc}   & 2.17 & 2.91 & 5.76 & - & - & - & -  \\ 
& RAFT~\cite{teed2020raft}  & 1.95 & 2.57 &  4.23 & - & - & - & - \\
\hline
\multicolumn{1}{c}{\multirow{14}{*}{C+T}}           & HD3~\cite{yin2019hierarchical} &  3.84      & 8.77      & 13.17  & 24.0  & - & - & - \\
 & LiteFlowNet~\cite{hui2018liteflownet} & 2.48 & 4.04 & 10.39 & 28.5 & - & - & -  \\
 & PWC-Net~\cite{sun2018pwc} & 2.55 & 3.93 & 10.35 & 33.7 & - & - & - \\
 & LiteFlowNet2~\cite{hui2020lightweight} & 2.24 & 3.78 & 8.97 & 25.9 & - & - & - \\
 & S-Flow~\cite{zhang2021separable}  & 1.30 & 2.59 & 4.60 & 15.9 & & & \\
 & RAFT~\cite{teed2020raft}   & 1.43 & 2.71 & 5.04 & 17.4 & - & - & - \\
 & FM-RAFT~\cite{jiang2021learning2}   & 1.29 & 2.95 & 6.80 & 19.3 & - & - & -  \\
 & GMA~\cite{jiang2021learning}   & 1.30 & 2.74 & 4.69 & 17.1 & - & - & - \\
 & GMFlow~\cite{xu2022gmflow}   & 1.08 & 2.48 & - & - & - & - & - \\
 & GMFlowNet~\cite{zhao2022global}   & 1.14 & 2.71 & 4.24 & 15.4 & - & - & - \\
 & CRAFT~\cite{sui2022craft}   & 1.27 & 2.79 & 4.88 & 17.5 & - & - & - \\
 & SKFlow~\cite{sun2022skflow}   & 1.22 & 2.46 & 4.47 & 15.5 & - & - & - \\
 
%  & GMA-L~\cite{jiang2021learning}   & 1.33~(+33\%) & 2.56~(+4\%) & 4.40 & 15.93 & - & - & - & &  \\
%  & GMA-Twins~\cite{jiang2021learning}   & 1.15~(+15\%) & 2.73~(+11\%) & 4.98 & 16.82 & - & - & - & &  \\
 & FlowFormer (Ours)   & $\uline{0.94}^{\dag}$ & $\uline{2.33}^{\dag}$ & $\uline{4.09}^{\dag}$ & $\uline{14.72}^{\dag}$ & - & - & - \\
 & FlowFormer+MCVA (Ours) & $\mathbf{0.90}^{\dag}$ & $\mathbf{2.30}^{\dag}$ & $\mathbf{3.93}^{\dag}$ & $\mathbf{14.13}^{\dag}$ & - & - & - \\
 % & Ours (tile)   & $0.95$ & $2.35$ & $\mathbf{4.09}$ & $\mathbf{14.72}$ & - & - & -
% \\ 
\hline
\multirow{16}{*}{C+T+S+K+H}                         &    LiteFlowNet2~\cite{hui2020lightweight}    &    (1.30)                       &       (1.62)                    &   (1.47)                          &      (4.8)                      &       3.48                     &         4.69                   &          7.74   \\
 &  PWC-Net+~\cite{sun2019models} & (1.71) & (2.34) &  (1.50) &  (5.3) & 3.45 &  4.60 & 7.72\\
 & VCN~\cite{yang2019volumetric} & (1.66) & (2.24) & (1.16) & (4.1) & 2.81 & 4.40 & 6.30 \\
 &  MaskFlowNet~\cite{zhao2020maskflownet} & - & - & - & - & 2.52 & 4.17 & 6.10 \\
  &  S-Flow~\cite{zhang2021separable} & (0.69) & (1.10) & (0.69) & (1.60) & 1.50 & 2.67 & $\uline{4.64}$ \\
 &  RAFT~\cite{teed2020raft} & (0.76) & (1.22) & (0.63) & (1.5) & 1.94 & 3.18 & 5.10
\\
&  RAFT*~\cite{teed2020raft} & (0.77) & (1.27) & - & - & 1.61 & 2.86 & 5.10
\\
 &  FM-RAFT~\cite{jiang2021learning2} & (0.79) & (1.70) & (0.75) & (2.1) & 1.72 & 3.60 &  6.17 \\
 &  GMA~\cite{jiang2021learning} & - & - & - & - & 1.40 & 2.88 & 5.15 \\
  &  GMA*~\cite{jiang2021learning} & (0.62) & (1.06) & (0.57) & (1.2) & 1.39 & 2.47 & 5.15 \\
  &  GMFlow~\cite{xu2022gmflow} & - & - & - & - & 1.74 & 2.90 & 9.32 \\
  &  GMFlowNet~\cite{zhao2022global} & (0.59) & (0.91) & (0.64) & (1.51) & 1.39 & 2.65 & 4.79 \\
  &  CRAFT~\cite{sui2022craft} & (0.60) & (1.06) & (0.57) & (1.20) & 1.45 & 2.42 & 4.79 \\
  &  SKFlow*~\cite{sun2022skflow} & (0.52) & (0.78) & (0.51) & (0.94) & 1.28 & 2.23 & 4.84 \\
&  FlowFormer (Ours) & (0.48) & (0.74) & (0.53) & (1.11) & $\uline{1.16}$ & $\uline{2.09}$ & $4.68^\dag$  \\
&  FlowFormer+MCVA (Ours) & (0.40) & (0.60) & (0.57) & (1.16) & $\mathbf{1.07}$ & $\mathbf{1.94}$ & $\mathbf{4.52}{^\dag}$  \\
\hline
\end{tabular}
% }
\end{center}
\caption{\label{Tab: comparison} $\textbf{Experiments on Sintel~\cite{butler2012naturalistic} and KITTI~\cite{geiger2013vision} datasets.}$ `A' denotes the autoflow dataset. `C + T' denotes training only on the FlyingChairs and FlyingThings datasets. `+ S + K + H' denotes finetuning on the combination of Sintel, KITTI, and HD1K training sets.  * denotes that the methods use the warm-start strategy~\cite{teed2020raft}, which relies on previous image frames in a video, while other methods use two frames only. $^\dag$ denotes the result is obtained via the tile technique.
Our FlowFormer+MCVA achieves the best generalization performance~(C+T) and ranks 1st on both the Sintel and the KITTI-15 benchmarks~(C+T+S+K+H).
}
\end{table*}

\subsection{Quantitative Comparison}

As shown in Tab.~\ref{Tab: comparison}, we evaluate FlowFormer and FlowFormer+MCVA on the Sintel and KITII-2015 benchmarks. Following previous methods, we evaluate the generalization performance of models on the training sets of Sintel and KITTI-2015 (denoted as `C+T'). We also compare the dataset-specific accuracy of optical flow models after dataset-specific finetuning (denoted as `C+T+S+K+H'). Autoflow~\cite{sun2021autoflow} is a synthetic dataset of complicated visual disturbance, while its training code is unreleased. 
% We copy the reported results from Autoflow~\cite{sun2021autoflow} for reference.a

\noindent \textbf{Generalization Performance.}
The `C+T' setting in Tab.~\ref{Tab: comparison} reflects the generalization capacity of models. 
FlowFormer outperforms all other methods except FlowFormer+MCVA, which reveals the superiority of the proposed transformer architecture.
FlowFormer+MCVA ranks 1st on both benchmarks among published methods. It achieves 0.90 and 2.30 on the clean and final pass of Sintel. Compared with FlowFormer, it further achieves 4.26\% error reduction on Sintel clean pass. On the KITTI-2015 training set, FlowFormer+MCVA achieves 3.93 F1-epe and 14.13 F1-all, improving FlowFormer by 0.16 and 0.59, respectively. These results show that our proposed MCVA promotes the generalization capacity of FlowFormer.

\noindent \textbf{Sintel Benchmark.}
After training the models in the `C+T+S+K+H' setting, we evaluate its performance on the Sintel online benchmark. 
FlowFormer sets new state-of-the-art performance on the Sintel benchmark, i.e. 1.16 on the clean pass and 2.09 on the final pass.
FlowFormer+MCVA further improves the performance and achieves 1.07 and 1.94 on the two passes.

% It achieves 1.07 and 2.09 on the clean and final passes, a 7.76\% and 7.18\% error reduction from previous best model FlowFormer.
\noindent \textbf{KITTI-2015 Benchmark.}
We further finetune our models on the KITTI-2015 training set after the Sintel stage and evaluate its performance on the KITTI online benchmark. 
FlowFormer achieves 4.68, ranking 2nd on the KITTI-2015 benchmark.
S-Flow~\cite{zhang2021separable} obtains slightly smaller error than FlowFormer on KITTI~($-$0.85\%), which, however, is significantly worse on Sintel~(31.6\% and 22.5\% larger error on clean and final pass).
S-Flow finds corresponding points by computing the coordinate expectation weighted by refined cost maps.
Images in the KITTI dataset are captured in urban traffic scenes, which contains objects that are mostly rigid.
Flows on rigid objects are rather simple, which is easier for cost-based coordinate expectation, but the assumption can be easily violated in non-rigid scenarios such as Sintel.
Even though,
FlowFormer+MCVA achieves 4.52 F1-all, improving FlowFormer by 0.16 while also outperforming the previous best model S-Flow by 0.12.

To conclude, FlowFormer outperforms all published works on all tracks except KITTI-2015 test due to the well-designed transformer architecture.
FlowFormer+MCVA further unleashes the FlowFormer capacity by enhancing the cost-volume encoder with the MCVA pretraining.

% MCVA-FlowFormer shows greater optical flow estimation capacity for both naturalistic non-rigid motions (Sinel) and real-world rigid scenarios (KITTI-2015). This validates that our proposed MCVA improves the FlowFormer architecture by enhancing the cost-volume encoder.

\subsection{Qualitative Comparison}

\begin{figure*}[t!]
    \centering
    \setlength\tabcolsep{1pt}
    \begin{tabular}{ccc}
        \includegraphics[width=0.33\linewidth]{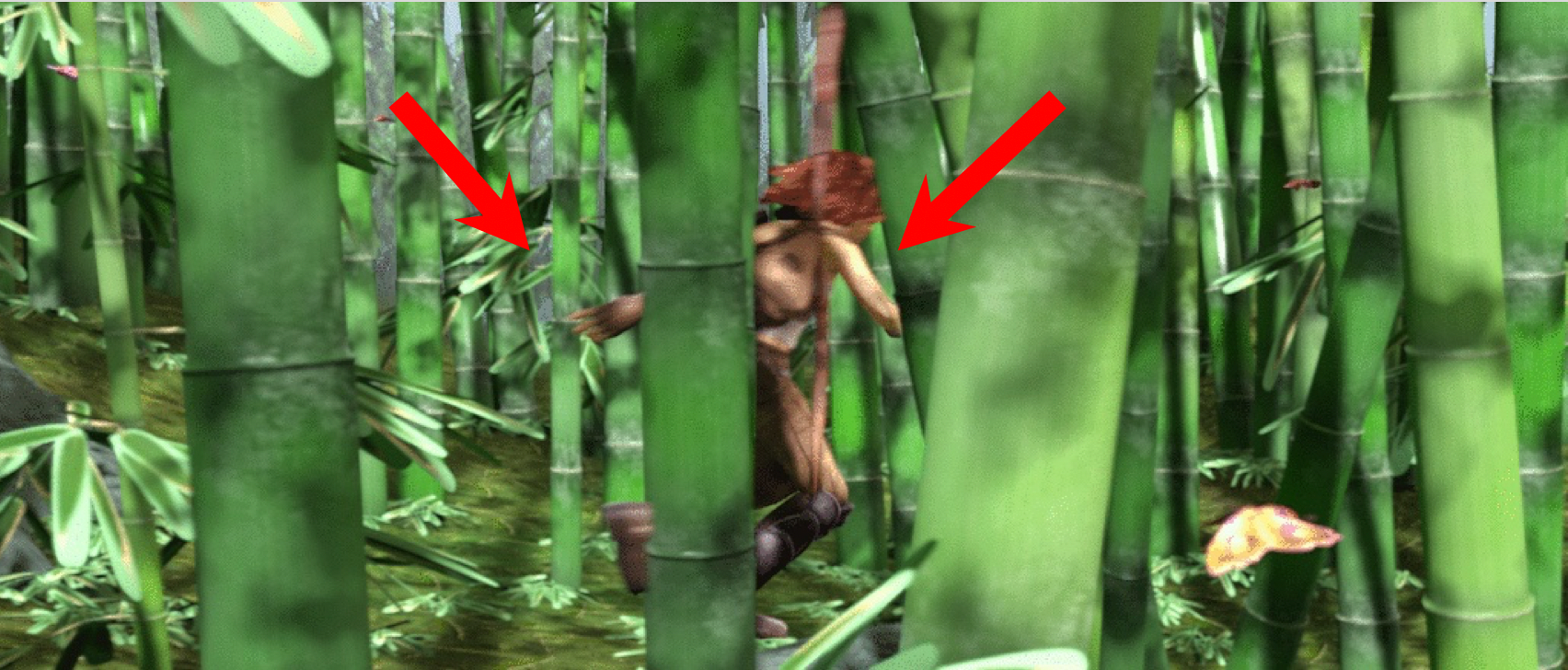}
        &
        \includegraphics[width=0.33\linewidth]{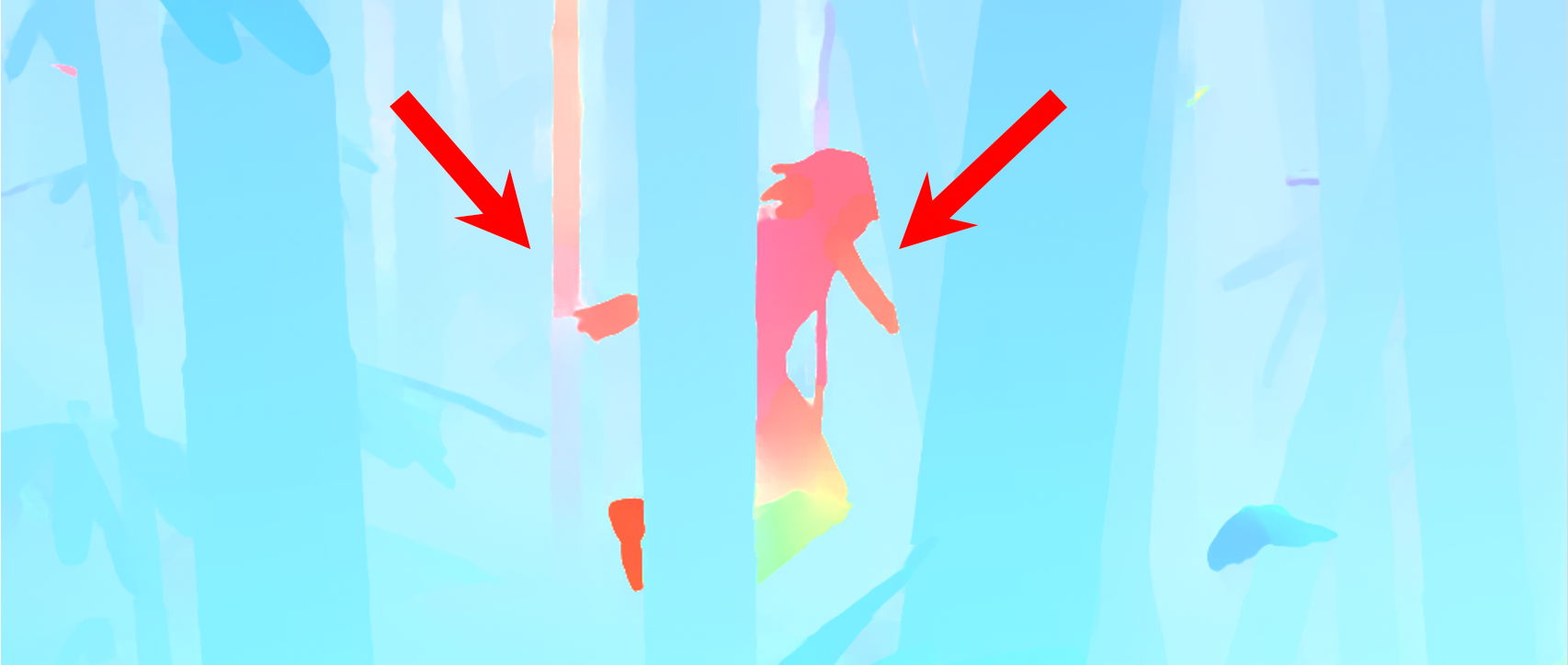}
        &
        \includegraphics[width=0.33\linewidth]{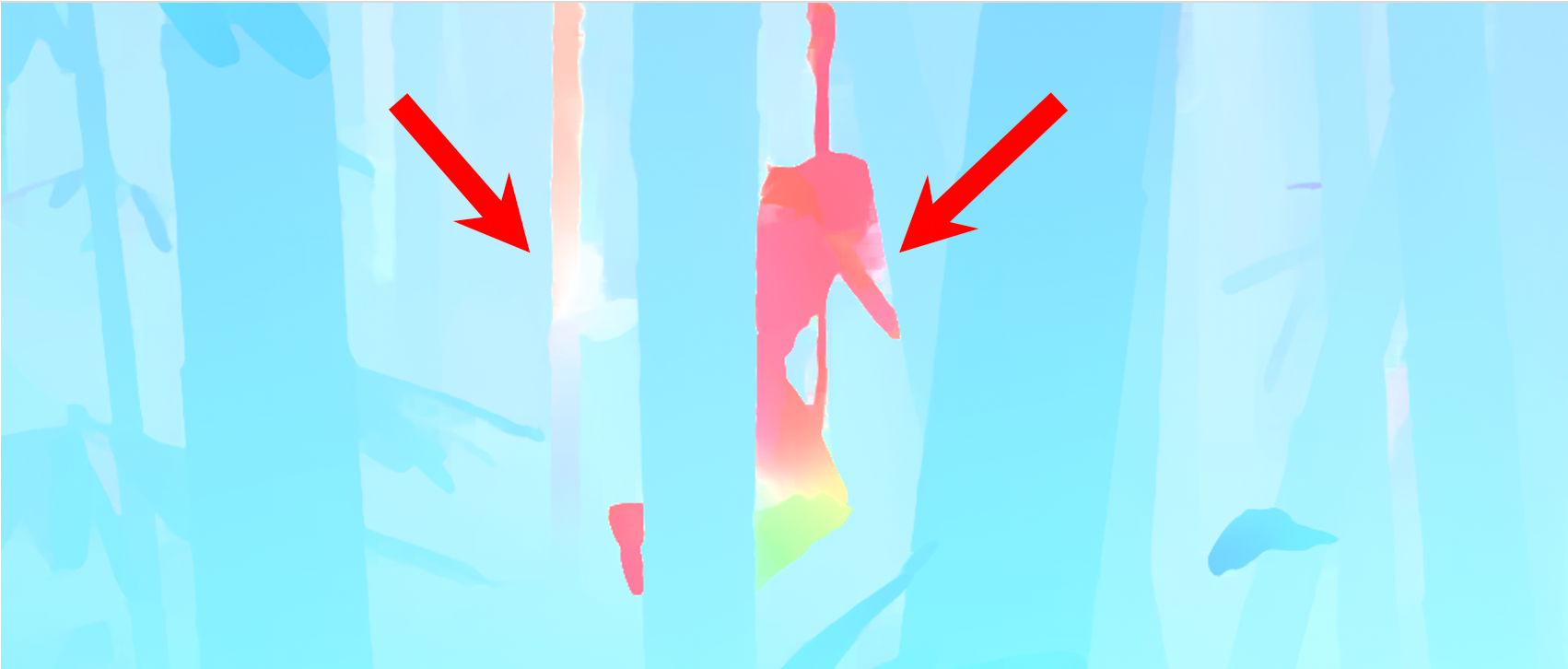}
        \\
        \includegraphics[width=0.33\linewidth]{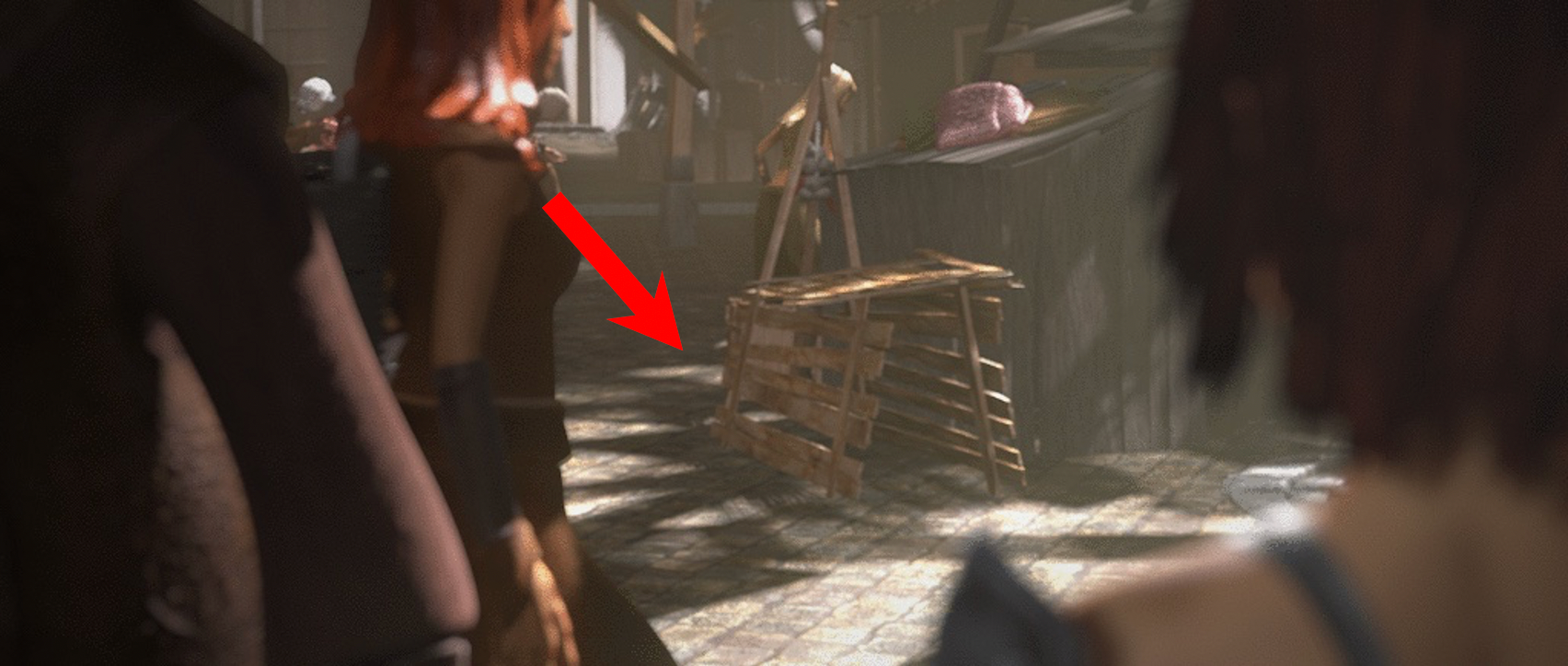} &
        \includegraphics[width=0.33\linewidth]{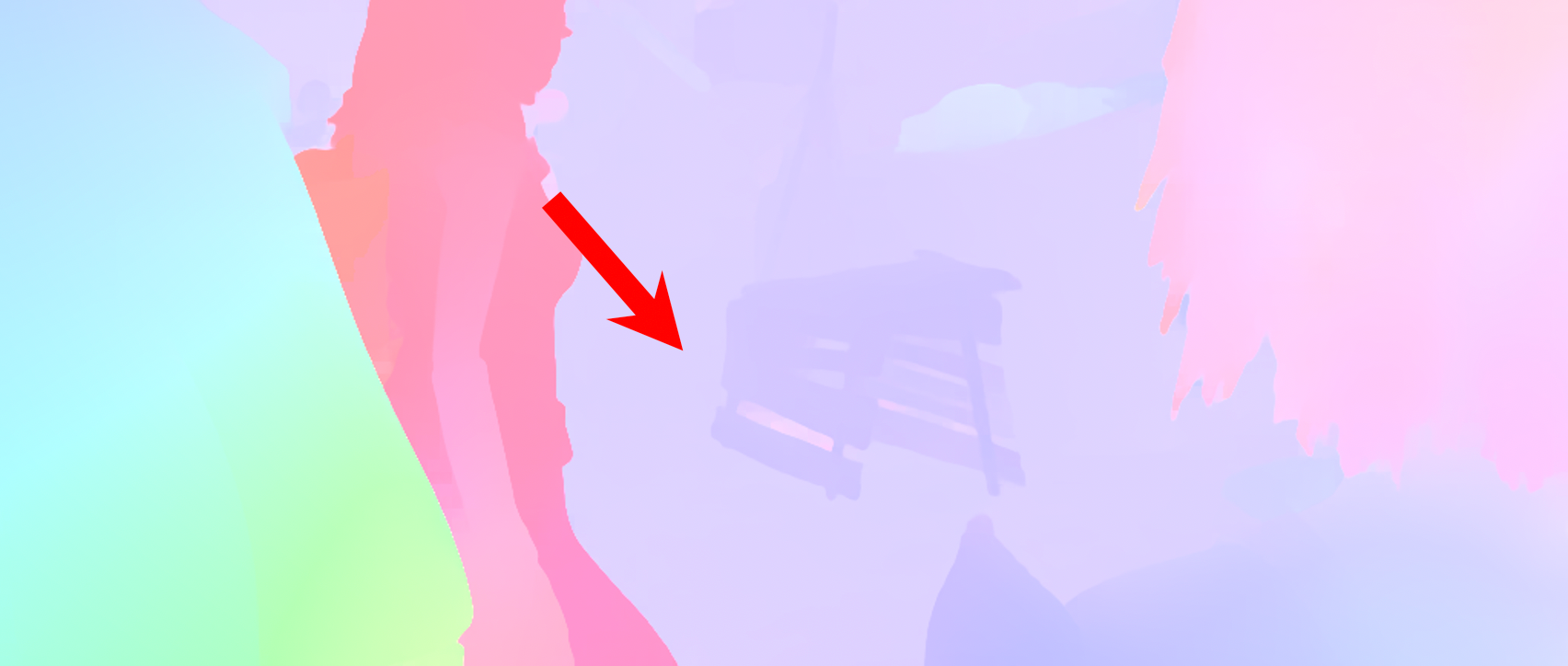} &
        \includegraphics[width=0.33\linewidth]{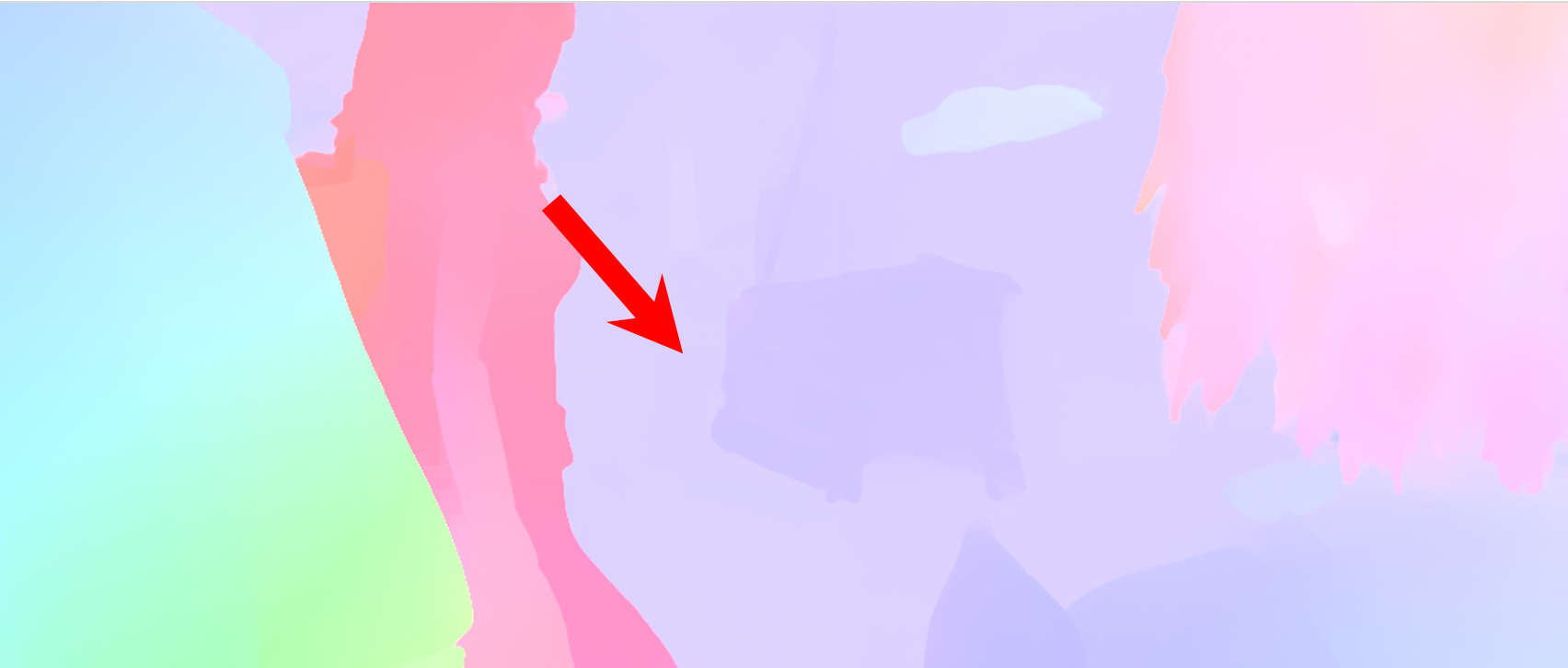} \\
         Input &  FlowFormer & GMA \\
        \includegraphics[width=.32\linewidth]{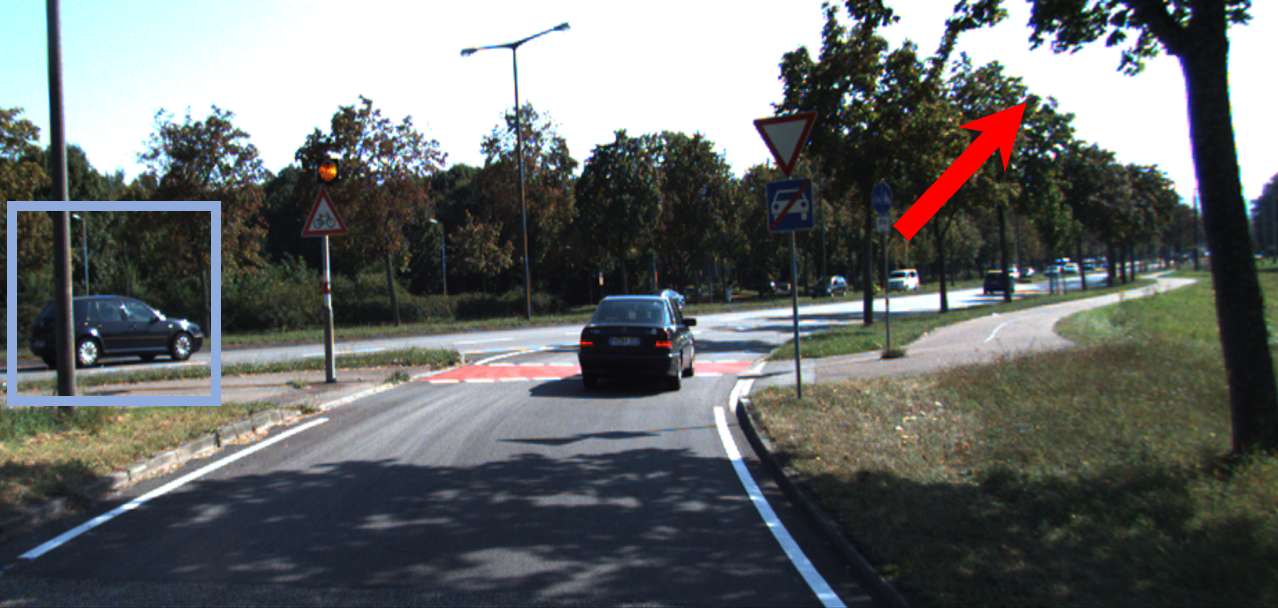} &
        \includegraphics[width=.32\linewidth]{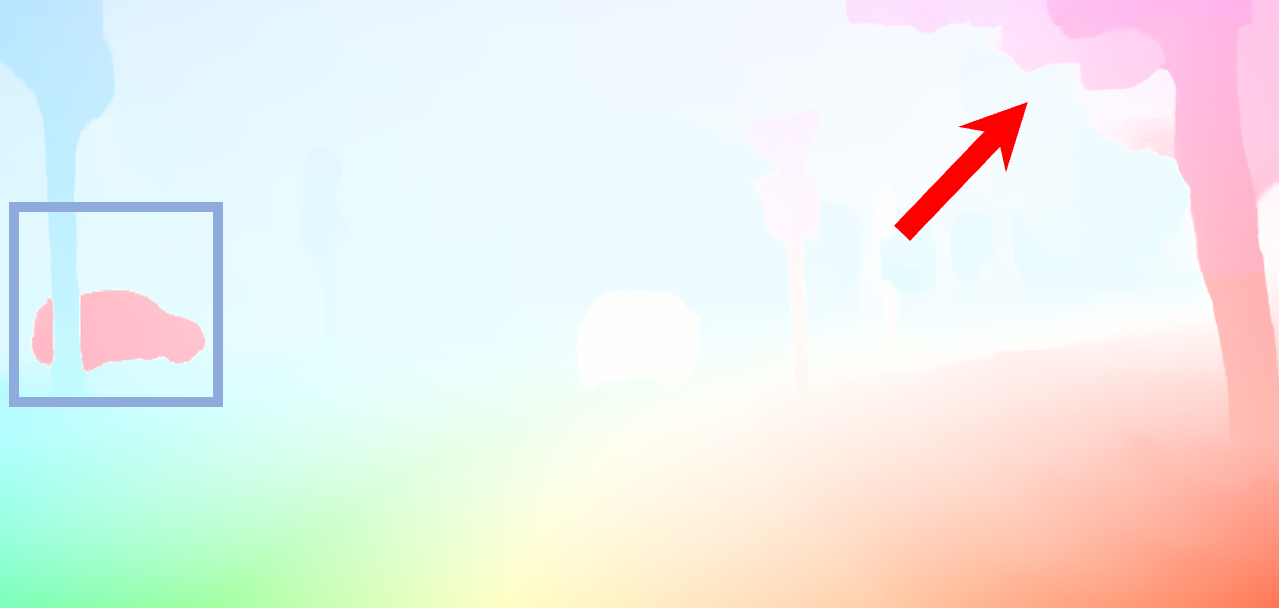} &
        \includegraphics[width=.32\linewidth]{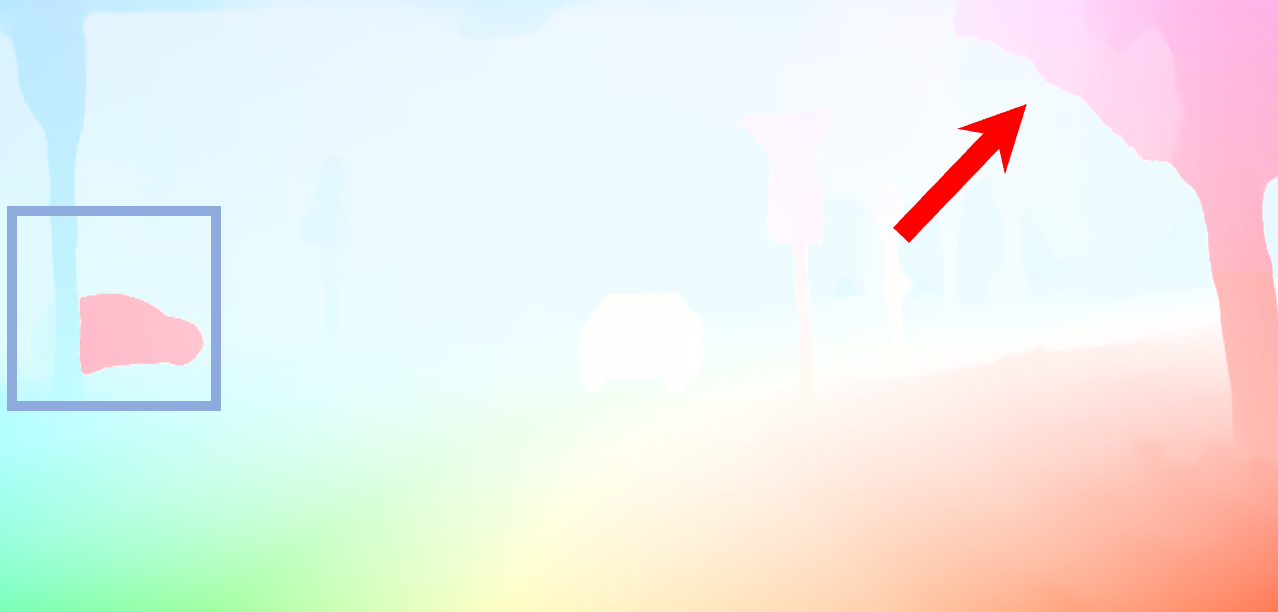} \\
        Input & FlowFormer+MCVA & FlowFormer \\
    \end{tabular}
    \caption{\textbf{Qualitative comparison.} As pointed by red arrows, FlowFormer greatly reduces the flow leakage around object boundaries and clearer details. FlowFormer+MCVA further improves the flow details and maintains better global consistency (indicated by the blue box).
    }
    \label{Fig: flow comparison}
\end{figure*}

% \begin{figure*}
% \centering
%     % \resizebox{1.0\linewidth}{!}{
% \setlength{\tabcolsep}{2pt}
% \begin{tabular}{@{} c c c @{}}
%         \includegraphics[width=.32\linewidth]{figures/example_flows/1.png} &
%         \includegraphics[width=.32\linewidth]{figures/example_flows/2.png} &
%         \includegraphics[width=.32\linewidth]{figures/example_flows/3.png}  \\
%         % A & B & C  \\
%         \includegraphics[width=.32\linewidth]{figures/example_flows/4.png} &
%         \includegraphics[width=.32\linewidth]{figures/example_flows/5.png} &
%         \includegraphics[width=.32\linewidth]{figures/example_flows/6.png} \\
%         Input & FlowFormer & FlowFormer+MCVA  \\
%     \end{tabular}
%     % }
%     \vspace{-0.15cm}
%     \caption{$\textbf{Qualitative comparison on Sintel and KITTI test sets.}$ FlowFormer+MCVA preserves clearer details (pointed by red arrows) and maintains better global consistency (indicated by blue boxes).
%     }
%     \label{fig:quality}
%     % \vspace{-0.3cm}
% \end{figure*}

\begin{table*}[t]
\centering
% \resizebox{1.0\linewidth}{!}{
\begin{tabular}{clccccc}
\hline
\multicolumn{1}{c}{\multirow{2}{*}{Experiment}} & \multicolumn{1}{c}{\multirow{2}{*}{Method}} & \multicolumn{2}{c}{Sintel (train)}                    & \multicolumn{2}{c}{KITTI-15 (train)}                 & \multirow{2}{*}{Params.}  \\
\cmidrule(r{1.0ex}){3-4}\cmidrule(r{1.0ex}){5-6}
\multicolumn{1}{c}{}                               & \multicolumn{1}{c}{}                        & \multicolumn{1}{c}{Clean} & \multicolumn{1}{c}{Final} & \multicolumn{1}{c}{F1-epe} & \multicolumn{1}{c}{F1-all} &    \\ 
\hline
\hline
% \multicolumn{6}{l}{\textit{baseline: Encoder Depth:0, image encoder: CNN}}
% \\
baseline & RAFT  & 1.53 & 2.99 & 5.73 & 18.29 & 5.3M    \\ 
\hline
\multirow{3}{*}{MCR$\rightarrow$LCT+CMD}                       
% &    RAFT    & - & - & - & - & M    \\ 
& $K=4$, $D=32$ & 1.66 & 2.93 & 5.60 & 19.67 &  5.5M \\
& $K=8$, $D=32$   & 1.58 & 2.90 & 5.50 & 18.71 &  5.5M  \\ 
 & $K=8$, $D=128$ & 1.44 & 2.80 & 5.22 & 17.64 &  5.6M  \\ 
\hline
\multirow{3}{*}{CNN $\rightarrow$ Twins}
 & CNN & 1.44 & 2.80 & 5.22 & 17.64 &  5.6M  \\ & Twins from Scratch & 1.44 & 2.86 & 5.38 & 17.58 &  14.0M \\
& Pretrained Twins    & 1.29 & 2.72 & 4.82 & 16.16 &   14.0M  \\ 
\hline
\multirow{5}{*}{Cost Encoding}                        
& None    & 1.29 & 2.72 & 4.82 & 16.16 &   14.0M  \\ 
& +Intra. & 1.29 & 2.89 & 4.74 & 15.71 &  14.1M \\
& AGT$\times$1 (+Intra.+Inter.) & 1.20 & 2.85 & 4.57 & 15.46 & 15.2M  \\ 
& AGT$\times$2 & 1.16 & 2.66 & 4.70 & 16.01 & 16.4M  \\ 
& AGT$\times$3 & 1.10 & 2.57 & 4.45 & 15.15 & 17.6M  \\ 
% \hline
% {RAFT $\xrightarrow[]{}$ GMA}                         &   GMA-ConvGRU    & - & - & - & - &   18.2M  \\ 
% & No & - & - & - & - &  15.1M \\
\hline
\end{tabular}
% }
\caption{\label{Tab: ablation} \textbf{Ablation study on the FlowFormer architecture}. We gradually change one component of the RAFT at a time to obtain our FlowFormer model. MCR$\rightarrow$LCT+CMD: replacing RAFT's decoder with OUR latent cost tokens + cost memory decoder. CNN$\rightarrow$Twins: replacing RAFT's CNN encoder with Twins-SVT transformer. Cost Encoding: adding intra-cost-map and inter-cost-map to form an Alternate-Group Transformer layer in the encoder. 3 AGT layers are used in our final model.
%In each experiment, we use the best model in the last experiment as the baseline. 
%In the \textit{MCR$\rightarrow$LCT+CMD} experiment, we replace the MCR used by RAFT with our proposed LCT+CMD.
%In the \textit{Cost Encoding} experiment, we first add the intra-cost-map attention~(+Intra.) and then add the inter-cost-map attention~(+Intra.+Inter.), which constitutes the complete AGT layer. Then, we increase the number of AGT layers to 3.
}
% \vspace{-15pt}
\end{table*}

\noindent \textbf{GMA v.s. FlowFormer}
We visualize flows that estimated by our FlowFormer and GMA of three examples in Fig.~\ref{Fig: flow comparison} to qualitatively show how FlowFormer outperforms GMA.
As transformers can encode the cost information at a large perceptive field, FlowFormer can distinguish overlapping objects via contextual information and thus reduce the leakage of flows over boundaries.
Compared with GMA, the flows that are estimatd by FlowFormer on boundaries of the bamboo and the human body are more precise and clear.

\noindent \textbf{FlowFormer v.s. FlowFormer+MCVA}
MCVA pretraining encourages the cost-volume encoder to aggregate cost information in a long range.
We visualize flow predictions by our FlowFormer and FlowFormer+MCVA on KITTI test sets in Fig.~\ref{Fig: flow comparison} to qualitatively show how MCVA pretraining improves FlowFormer. 
The red arrows highlight that FlowFormer+MCVA preserves clearer details than FlowFormer and shows greater global aggregation capacity indicated by blue boxes. 

\subsection{Ablation Study on FlowFormer Architecture}

\begin{table*}[h]
\centering
% \resizebox{1.0\linewidth}{!}{
\begin{tabular}{lccccccc}
% \multicolumn{1}{c}{\multirow{2}{*}{Training Data}} 
& \multicolumn{1}{c}{\multirow{2}{*}{Method}} & \multicolumn{2}{c}{Sintel (train)}                    & \multicolumn{2}{c}{KITTI-15 (train)} & \multirow{2}{*}{parameters} \\
\cmidrule(r{1.0ex}){3-4}\cmidrule(r{1.0ex}){5-6}
\multicolumn{1}{c}{}                               & \multicolumn{1}{c}{}                        & \multicolumn{1}{c}{Clean} & \multicolumn{1}{c}{Final} & \multicolumn{1}{c}{F1-epe} & \multicolumn{1}{c}{F1-all} & \\ 
\hline
% \multirow{4}{*}{C+T} 
 & GMA~\cite{jiang2021learning}   & 1.30~(+30\%) & 2.74~(+12\%) & 4.69~(+15\%) & 17.1~(+16\%) & 5.9M \\
 & FlowFormer~(small)  & 1.20~(+20\%) & 2.64~(+8\%) & 4.57~(+12\%) & 16.62~(+13\%) & 6.2M \\
 \cline{2-7}
% \cmidrule(r{1.0ex}){2-7}
 & GMA-L~\cite{jiang2021learning}   & 1.33~(+33\%) & 2.56~(+4\%) & 4.40~(+8\%) & 15.93~(+8\%) & 17.0M \\
 & GMA-Twins~\cite{jiang2021learning}   & 1.15~(+15\%) & 2.73~(+11\%) & 4.98~(+22\%) & 16.82~(+14\%) & 14.2M \\
 & FlowFormer   & $\mathbf{1.00}$ & $\mathbf{2.45}$ & $\mathbf{4.09}$ & $\mathbf{14.72}$ & 18.2M
\\ \hline
\end{tabular}
% }
\caption{\label{Tab: param comparison} 
\textbf{FlowFormer v.s. GMA architectures}. 
Ours~(small) is a small version of FlowFormer and uses the CNN image feature encoder of GMA.
GMA-L is a large version of GMA. 
GMA-Twins replace its CNN image feature encoder with pre-trained Twins.
(+x\%) indicates that this model obtains x\% larger error than ours.
}
\end{table*}

We conduct a series of ablation experiments in Tab.~\ref{Tab: ablation}.
We start from RAFT as the baseline, which directly regresses residual flows with the multi-level cost retrieval~(MCR) decoder,
and gradually replace its components with our proposed components.
We first replace RAFT's MCR decoder with the latent cost tokenization (LCT) part of our encoder and the cost memory decoder (CMD) (denoted as `MCR$\rightarrow$LCT+CMD'). Note that our cost memory decoder cannot be used alone on top of the 4D cost volume of RAFT because of the too large number of tokens. It must be combined with our latent cost tokens ($\bf T_x$ from Eq. (1)). Encoding $K=8$ latent tokens of $D=128$ dimensions for each source pixel achieves the best performance.
%latent cost tokens~(LCT), which tokenizes each cost map into $K$ tokens of length $D$ and regresses residual flows via cross attention with our cost memory decoder~(denoted as `MCR $\xrightarrow[]{}$ LCT').
Based on LCT+CMD with $K=8$ and $D=128$, we further replace RAFT's CNN image feature encoder with Twins-SVT~(denoted as `CNN$\rightarrow$Twins').
We then further add attention layers of the proposed cost-volume encoder to encode and update latent cost tokens.
The proposed Alternate-Group Transformer (AGT) layer consists of two types of attention, i.e., intra-cost-map attention and inter-cost-map attention.
We first add a single intra-cost-map attention layer (denoted as `+Intra.'), and then add the inter-cost-map attention (denoted as `AGT$\times1$ (+Intra.+Inter.)', which is equivalent to adding a single AGT layer.
We then test on increasing the number of AGT layers to 2 and 3.
% Finally, we switch the RAFT-ConvGRU module to GMA-ConvGRU module, which is the final version of our FlowFormer.
Following RAFT, all models are trained on FlyingChairs~\cite{dosovitskiy2015flownet} with 100k iterations and FlyingThings~\cite{mayer2016large} with 60k iterations, and then evaluated on the training set of Sintel~\cite{butler2012naturalistic} and KITTI-2015~\cite{geiger2013vision}.

\noindent \textbf{MCR $\rightarrow$ LCT+MCD.}
The number of latent tokens $K$ and token dimension $D$ determine how much cost volume information the cost tokens can encode.
From $K=4, D=32$ to $K=8, D=128$, the AEPE decreases because the cost tokens summarizes more cost map information and benefits the residual flow regression.
The latent cost tokens are capable of summarizing whole-image information and our MCD can absorb interested information from them through co-attention, while the MCR decoder of RAFT only retrieves multi-level costs inside flow-guided local windows.
Therefore, even without our AGT layers in our encoder,  LCT+MCD still shows better performance than MCR decoder of RAFT.
% In contrast to MCR retrieving multi-level costs inside flow-guided local windows, 

% Besides, event without cost token encoding, LCT regresses residual flow from cost tokens that summarizes the whole cost map while

\noindent \textbf{CNN vs. Transformer Image Encoder.}
In the CNN$\rightarrow$Twins experiment, 
the AEPE of Twins trained from scratch is marginally worse than CNN, but the ImageNet-pretraining is beneficial,
% the AEPE of FlowFormer persistently decreases from CNN to Twins and then to Pretrained Twins, which shows that the Twins is better than CNN as an image feature encoder and the ImageNet-pretraining is beneficial,
because Twins is a transformer architecture with larger receptive field and model capacity, which requires more training examples for sufficient training. %a more broad perceptive field and the pretraining makes up for the lack of training data to a certain extent.
% In Tab.~\ref{}, we notice that larger CNN and pre-trained twins also improve GMA but may lead to worse AEPE on some metrics, i.e. GMA v.s. GMA-L on the Sintel clean pass and GMA v.s. GMA-Twins on KITTI F1-epe, but improve FlowFormer in all metrics.
% We believe the reason is that the perceptive field is extended to the whole cost map in the decoder of FlowFormer.

% in RAFT with pre-trained Twins~(GMA-Twins). Compared with the original GMA~(GMA-CNN), GMA-Twins is better on Sintel clean but worse on other metrics, which does not present significant difference.
% We also replace the pre-trained Twins-SVT in FlowFormer with the CNN used by GMA~(Our-CNN).
% Our-CNN maintains an inferior performance compared to Ours-Twins, which denotes that the pre-trained Twins-SVT benefits our FlowFormer.
% Our-Twins achieves lower error on all of the four metrics than GMA-Twins. Such improvement comes from our proposed cost-volume encoder and decoder.

\noindent \textbf{Cost Encoding.} 
In the cost-volume encoder, we encode and update the latent cost tokens with an intra-cost-map attention operation and an inter-cost-map attention operation. The two operations form an Alternate-Group Transformer (AGT) layer.
Then we gradually increase the number of AGT layers to 3.
From no attention layer to AGT$\times$3, the errors gradually decrease, which demonstrates that encoding latent cost tokens with our AGT layers benefits flow estimation.

\begin{table}[t]
\centering
% \resizebox{1.0\linewidth}{!}{
\begin{tabular}{llllllll}
\hline
  & \multirow{2}{*}{Intra.} & \multirow{2}{*}{Inter.} & \multicolumn{2}{c}{Sinte (train)} & \multicolumn{2}{c}{Kitti (train)} & \multirow{2}{*}{Params.} \\
  &  &  & Clean & Final & F1-epe & F1-all & \\ 
\hline
 & Trans & Trans & \textbf{1.20} & 2.85 & \textbf{4.57} & \textbf{15.46} & 15.2M \\
 \hline
 & MLP & Trans & 1.20 & \textbf{2.67} & 5.01 & 16.81 & 15.2M \\
& Trans & Conv & 1.23 & 2.72 & 4.73 & 15.87 & 15.1M \\
& MLP & Conv & 1.22 & 2.71 & 4.88 & 17.23 & 15.1M \\
\hline
\end{tabular}
% }
\caption{\label{Tab: ablation transformer v.s. mlp} \textbf{Transformer v.s. MLP\&Conv in AGT}. For intra-cost-map aggregation layer~(Intra.), we replace transformer (Trans) with MLP-Mixer~\cite{tolstikhin2021mlpmixer} block (MLP). For inter-cost-map aggregation layer~(Inter.), we replace transformer with ConvNeXt~\cite{Liu2022ACF} block (Conv).
}
\end{table}

\begin{table}[t!]
\centering
% \resizebox{1.0\linewidth}{!}{
\begin{tabular}{lccccccc}
\hline
 \multirow{2}{*}{Sintel (train)} & \multicolumn{2}{c}{clean} & \multicolumn{2}{c}{final} \\
  & w/ tile & w/o tile & with tile & w/o tile \\
 \hline
 FlowFormer & \textbf{0.94} & 1.01 & \textbf{2.33} & 2.40 \\
\hline
\end{tabular}
% }
\caption{\label{Tab: gaussian tile} \textbf{FlowFormer with v.s. without tile}. We tile test images with the training image size. Tiling slightly improves performance.
}
\end{table}

\begin{table}[t!]
\centering
\resizebox{1.0\linewidth}{!}{
\begin{tabular}{lcccccc}
 \multicolumn{1}{c}{\multirow{2}{*}{Method}}                    & \multicolumn{2}{c}{Sintel}                     & \multicolumn{2}{c}{KITTI-15}   \\
\cmidrule(r{1.0ex}){2-3}\cmidrule(r{1.0ex}){4-5}                  & \multicolumn{1}{c}{Clean$\downarrow$} & \multicolumn{1}{c}{Final$\downarrow$} & \multicolumn{1}{c}{F1-epe$\downarrow$} &
 F1-all$\downarrow$\\ 
 \hline 
 FlowFormer  & $\mathbf{1.10}$ & $\mathbf{2.57}$ & $\mathbf{4.45}$ & \textbf{15.15}  \\
 w/o AGT PE  & $1.18$ & $2.63$ & $4.78$ & 16.19  \\
 w/o token PE & $1.19$ & $2.59$ & $4.63$ & 16.23 \\
\hline
\end{tabular}
}
\caption{\label{Tab: pe ablation} \textbf{FlowFormer with v.s. without positional encoding (PE)}. Removing PE at either the AGT stage or the tokenization stage
hampers FlowFormer.
}
\end{table}

\noindent \textbf{Transformer v.s. MLP\&Conv.} As shown in Table \ref{Tab: ablation transformer v.s. mlp}, we conduct ablation experiments on the alternative-group transformer (AGT) layer. For intra-cost-map aggregation layer, since the number and dimension of latent cost tokens are fixed, we test on replacing our design with MLP-Mixer~\cite{tolstikhin2021mlpmixer}~(2nd row), which is a state-of-the-art MLP-based architecture. We also substitute  ConvNeXt~\cite{Liu2022ACF} for transformer in inter-cost-map aggregation~(3rd row).
Furthermore, we replace both transformers with MLP and ConvNext~(4th row).
Replacing transformer layers leads to slightly better performance on Sintel final pass, while brings a clear drop on KITTI.
Therefore, we adopt the proposed full transformer architecture as our final model.

\begin{table}[!t]
\centering
% \scriptsize
% \tiny
% \small
% \resizebox{0.9\linewidth}{!}{
\begin{tabular}{ccccc}
\hline
 \multicolumn{1}{c}{\multirow{2}{*}{MLP Depth}} & \multicolumn{2}{c}{Sintel (train)}                    & \multicolumn{2}{c}{KITTI-15 (train)} \\
% \cmidrule(r{1.0ex}){3-4}\cmidrule(r{1.0ex}){5-6}\cmidrule(r{1.0ex}){7-8} \cmidrule(r{1.0ex}){9-9} 
\cmidrule(r{1.0ex}){2-3} \cmidrule(r{1.0ex}){4-5}
        & \multicolumn{1}{c}{Clean} & \multicolumn{1}{c}{Final} & \multicolumn{1}{c}{F1-epe} & \multicolumn{1}{c}{F1-all} \\ 
\hline
1 & $\underline{0.92}$ & $\mathbf{2.30}$ & 4.10 & 14.66 \\
3 & $\mathbf {0.90}$ & $\mathbf{2.30}$ & $\mathbf{3.93}$ & $\mathbf{14.13}$ \\ 
5 & 0.94 & $\underline{2.35}$ & $\underline{4.08}$ & $\underline{14.64}$ \\ 
\hline
\end{tabular}
% }
% \vspace{-0.2cm}
\caption{\label{Tab: mlp-depth} $\textbf{Reconstruction head depth.}$ We use an MLP as the reconstruction head during MCVA pretraining. The three layer MLP yields the best result.}
% \vspace{-0.2cm}
\end{table}

\begin{table}[t]
\centering
% \scriptsize
% \tiny
% \small
% \resizebox{0.9\linewidth}{!}{
\begin{tabular}{cccccc}
\hline
 %{\multirow{2}{*}{freeze image encoder}} 
 %&{\multirow{2}{*}{freeze context encoder}} 
  \multicolumn{2}{c}{Masking} 
 & \multicolumn{2}{c}{Sintel (train)}                    
 & \multicolumn{2}{c}{KITTI-15 (train)} \\
% \cmidrule(r{1.0ex}){3-4}\cmidrule(r{1.0ex}){5-6}\cmidrule(r{1.0ex}){7-8} \cmidrule(r{1.0ex}){9-9} 
\cmidrule(r{1.0ex}){1-2} \cmidrule(r{1.0ex}){3-4} \cmidrule(r{1.0ex}){5-6}
   Strategy  & Ratio & \multicolumn{1}{c}{Clean} & \multicolumn{1}{c}{Final} & \multicolumn{1}{c}{F1-epe} & \multicolumn{1}{c}{F1-all} \\ 
\hline
Block-sharing & 20\% & 0.97 & 2.41 & \textbf{3.91} & 14.27 \\
Block-sharing & 50\% & \textbf{0.90} & \textbf{2.30} & 3.93 & 14.13 \\ 
Block-sharing & 80\% & 0.94 & 2.35 & 4.00 & \textbf{14.07} \\ 
Random & 20\% & 0.98 & 2.39 & 4.12 & 14.89 \\ 
Random & 50\% & 0.93 & 2.35 & 4.03 & 14.30 \\
Random & 80\% & 0.92 & 2.38 & 4.10 & 14.51 \\
\hline
\end{tabular}
% }
% \vspace{-10pt}
\caption{\label{Tab: masking.} \textbf{Cost Volume Masking Strategy.} Masked cost volume autoencoding with different masking strategies and ratios. Block-sharing masking brings greater performance gain than random masking. Masking 50\% cost values yields the best overall results.}
% \vspace{-19pt}
\end{table}

\begin{table}[t]
\centering
% \scriptsize
% \tiny
% \small
% \resizebox{0.9\linewidth}{!}{
\begin{tabular}{lccccc}
\hline
 {\multirow{2}{*}{Location}} &{\multirow{2}{*}{Query Feature}} 
 & \multicolumn{2}{c}{Sintel (train)}                    
 & \multicolumn{2}{c}{KITTI-15 (train)} \\
% \cmidrule(r{1.0ex}){3-4}\cmidrule(r{1.0ex}){5-6}\cmidrule(r{1.0ex}){7-8} \cmidrule(r{1.0ex}){9-9} 
\cmidrule(r{1.0ex}){3-4} \cmidrule(r{1.0ex}){5-6}
    & \multicolumn{1}{c}{}    & \multicolumn{1}{c}{Clean} & \multicolumn{1}{c}{Final} & \multicolumn{1}{c}{F1-epe} & \multicolumn{1}{c}{F1-all} \\ 
\hline
Fixed & PE & 0.99 & $\uline{2.40}$ & 4.35 & 15.33 \\
Random & PE & $\uline{0.95}$ & 2.42 & $\uline{3.99}$ & $\uline{14.47}$ \\ 
Random & PE + Cropped patch & $\mathbf{0.90}$ & $\mathbf{2.30}$ & $\mathbf{3.93}$ & $\mathbf{14.13}$ \\ 
\hline
\end{tabular}
% }
\caption{\label{Tab: Pre-text reconstruction design.} $\textbf{Pre-text reconstruction design.}$ Our pre-text reconstruction task leads to better performance over conventional MAE task (first row) and its improved version with random reconstruction locations (second row).
}
\end{table}

\noindent \textbf{Optical flow tiling.} 
Transformers are sensitive to image size due to the use of positional encodings. We use the Gaussian tile technique to avoid image size misalignment between training and test stages. 
Tab.~\ref{Tab: gaussian tile} shows that the tiling technique slightly improves performance.

\noindent \textbf{FlowFormer with v.s. without positional encoding (PE).} Transformers encode tokens with PE to obtain position information. We add PE to cost tokens in two stages: the tokenization stage and the AGT cost encoding stage. Tab.~\ref{Tab: pe ablation} reveals PE is essential in both stages. Removing either PE leads to a performance drop.

\noindent \textbf{FlowFormer vs. GMA.} The full version of FlowFormer has 18.2M parameters, which is larger than GMA. One of the causes is that FlowFormer uses the first two stages of ImageNet-pretrained Twins-SVT as the image feature encoder while GMA uses a CNN.
We present an experiment to compare FlowFormer and GMA with aligned settings in Tab.~\ref{Tab: param comparison}.
We first provide a small version of FlowFormer using GMA's CNN image encoder and also set $K=4$, $D=32$, and AGT$\times$1.
Although the smaller version of FlowFormer (denoted as 'Ours (small)') has a significant performance drop compared to the full version of FlowFormer, it still outperforms GMA in terms of all metrics.
We also design two enhanced GMA models and compare them with the full version of FlowFormer to show that the performance improvements are not simply derived from adding more parameters.
The first one is denoted as `GMA-L', a large version of GMA and the second one is denoted as `GMA-Twins' which also adopts the pretrained Twins as the image encoder. 
%that encodes the image features with pre-trained Twins.
% To be more fair, we provide a small version FlowFormer, a larger versoin GMA~(GMA-L), and GMA with pre-trained Twins image feature encoder~(GMA-Twins) in Tab.~\ref{Tab: param comparison}, which can be compared under similar amount of parameters and the same image feature encoder.
% Ours~(small) exploits the CNN image feature encoder used by GMA, uses 4 cost tokens with length of 32, and encodes the cost tokens with 1 AGT layer~(\textit{K=4, D=32, Encoder Depth=1}).
In this experiment, we train all models on FlyingChairs with 120k iterations and FlyingThings with 120k iterations.
Similar to reducing RAFT to RAFT~(small)~\cite{teed2020raft}, GMA-L enlarges GMA by doubling feature channels, which has 17M parameters, comparable to FlowFormer.
However, its performance degrades in Sintel clean, a 33\% larger error than FlowFormer.
GMA-Twins replaces the CNN image encoder with the shallow Image-Net pre-trained Twins-SVT as FlowFormer does.
The largest improvement of GMA-Twins upon GMA is on the Sintel clean, but it still has a 15\% larger error than FlowFormer. GMA-Twins does not lead to significant error reduction on other metrics and is even worse on the KITTI-15 F1-epe.
In conclusion, the performance improvement of FlowFormer is not derived from more parameters but the novel design of the architecture.

\subsection{Ablation Study on MCVA Pretraining}

We conduct a set of ablation studies to show the effectiveness of designs in the Masked Cost-volume Autoencoding (MCVA). All models in the experiemnts are first pretrained and then finetuned on `C+T'. We report the test results on Sintel and KITTI training sets.

\begin{table}[t]
\centering
% \scriptsize
% \tiny
% \small
% \resizebox{0.9\linewidth}{!}{
\begin{tabular}{lcccc}
\hline
 \multirow{2}{*}{Size of ${\bf q}_{\bf x}^L$} & \multicolumn{2}{c}{Sintel (train)}                    & \multicolumn{2}{c}{KITTI-15 (train)} \\
% \cmidrule(r{1.0ex}){3-4}\cmidrule(r{1.0ex}){5-6}\cmidrule(r{1.0ex}){7-8} \cmidrule(r{1.0ex}){9-9} 
\cmidrule(r{1.0ex}){2-3} \cmidrule(r{1.0ex}){4-5}
        & \multicolumn{1}{c}{Clean} & \multicolumn{1}{c}{Final} & \multicolumn{1}{c}{F1-epe} & \multicolumn{1}{c}{F1-all} \\ \hline
$15 \times 15$ & $\mathbf{0.90}$ & $\underline{2.30}$ & $\underline{3.93}$  & $\mathbf{14.13}$ \\
$25 \times 25$ & $\underline{0.92}$ & $\mathbf{2.28}$ & $\mathbf{3.87}$ & $\underline{14.54}$ \\ 
\hline
\end{tabular}
% }
% \vspace{-10pt}
\caption{\label{Tab: size} 
\textbf{Size of ${\bf q}_{\bf x}^L$}. ${\bf q}_{\bf x}^L$ is the cost-map patch to be reconstructed in MVCA.
Further increasing the size of ${\bf q}_{\bf x}^L$ does not bring clear gain.
}
% \vspace{-9pt}
\end{table}

\begin{table}[t]
\centering
% \scriptsize
% \tiny
% \small
% \resizebox{0.9\linewidth}{!}{
\begin{tabular}{cccccc}
\hline
 %{\multirow{2}{*}{freeze image encoder}} 
 %&{\multirow{2}{*}{freeze context encoder}} 
  \multicolumn{2}{c}{Freeze} 
 & \multicolumn{2}{c}{Sintel (train)}                    
 & \multicolumn{2}{c}{KITTI-15 (train)} \\
% \cmidrule(r{1.0ex}){3-4}\cmidrule(r{1.0ex}){5-6}\cmidrule(r{1.0ex}){7-8} \cmidrule(r{1.0ex}){9-9} 
\cmidrule(r{1.0ex}){1-2} \cmidrule(r{1.0ex}){3-4} \cmidrule(r{1.0ex}){5-6}
  IE  & CE & \multicolumn{1}{c}{Clean} & \multicolumn{1}{c}{Final} & \multicolumn{1}{c}{F1-epe} & \multicolumn{1}{c}{F1-all} \\ 
\hline
\xmark & \xmark & - & - & - & - \\
\xmark & \cmark & - & - & - & - \\ 
\cmark & \xmark & $\uline{0.92}$ & $\uline{2.34}$ & $\mathbf{3.90}$ & $\uline{14.23}$ \\ 
\cmark & \cmark & $\mathbf{0.90}$ & $\mathbf{2.30}$ & $\uline{3.93}$ & $\mathbf{14.13}$ \\ 
\hline
\end{tabular}
% }
\caption{\label{Tab: freezing encoders.} \textbf{Freezing the image encoder (IE) and the context encoder (CE) in pretraining.} Freezing the Image-Net pretrained image encoder ensures the reconstruction targets (\textit{i.e.}, raw cost values) maintain static, otherwise the model diverges. Freezing the context encoder leads to better overall performance.
}
\end{table}

\begin{table}[t]
\centering
% \scriptsize
% \tiny
% \small
% \resizebox{0.9\linewidth}{!}{
\begin{tabular}{lcccc}
\hline
 \multicolumn{1}{c}{\multirow{2}{*}{Methods}} & \multicolumn{2}{c}{Sintel (train)}                    & \multicolumn{2}{c}{KITTI-15 (train)} \\
% \cmidrule(r{1.0ex}){3-4}\cmidrule(r{1.0ex}){5-6}\cmidrule(r{1.0ex}){7-8} \cmidrule(r{1.0ex}){9-9} 
\cmidrule(r{1.0ex}){2-3} \cmidrule(r{1.0ex}){4-5}
        & \multicolumn{1}{c}{Clean} & \multicolumn{1}{c}{Final} & \multicolumn{1}{c}{F1-epe} & \multicolumn{1}{c}{F1-all} \\ 
\hline
Unsupervised Baseline & $\uline{0.99}$ & $\uline{2.54}$ & $\uline{4.38}$ & $\uline{15.22}$ \\ 
MCVA (ours)  & $\mathbf {0.90}$ & $\mathbf {2.30}$ & $\mathbf {3.93}$ & $\mathbf{14.13}$  \\ 
\hline
\end{tabular}
% }
\caption{\label{Tab: unsupervised} $\textbf{Comparisons with unsupervised methods.}$ We use the conventional unsupervised algorithm~\cite{stone2021smurf,9477059,Luo_2021_CVPR} (using photometic loss and smooth loss) to pretrain FlowFormer for comparison. Our MCVA outperforms the unsupervised counterpart.
}
\end{table}

\begin{figure*}[t!]
    \centering
    \resizebox{0.95\linewidth}{!}{
        \includegraphics[width=0.98\linewidth, trim={0mm 0mm 0mm 0mm}, clip]{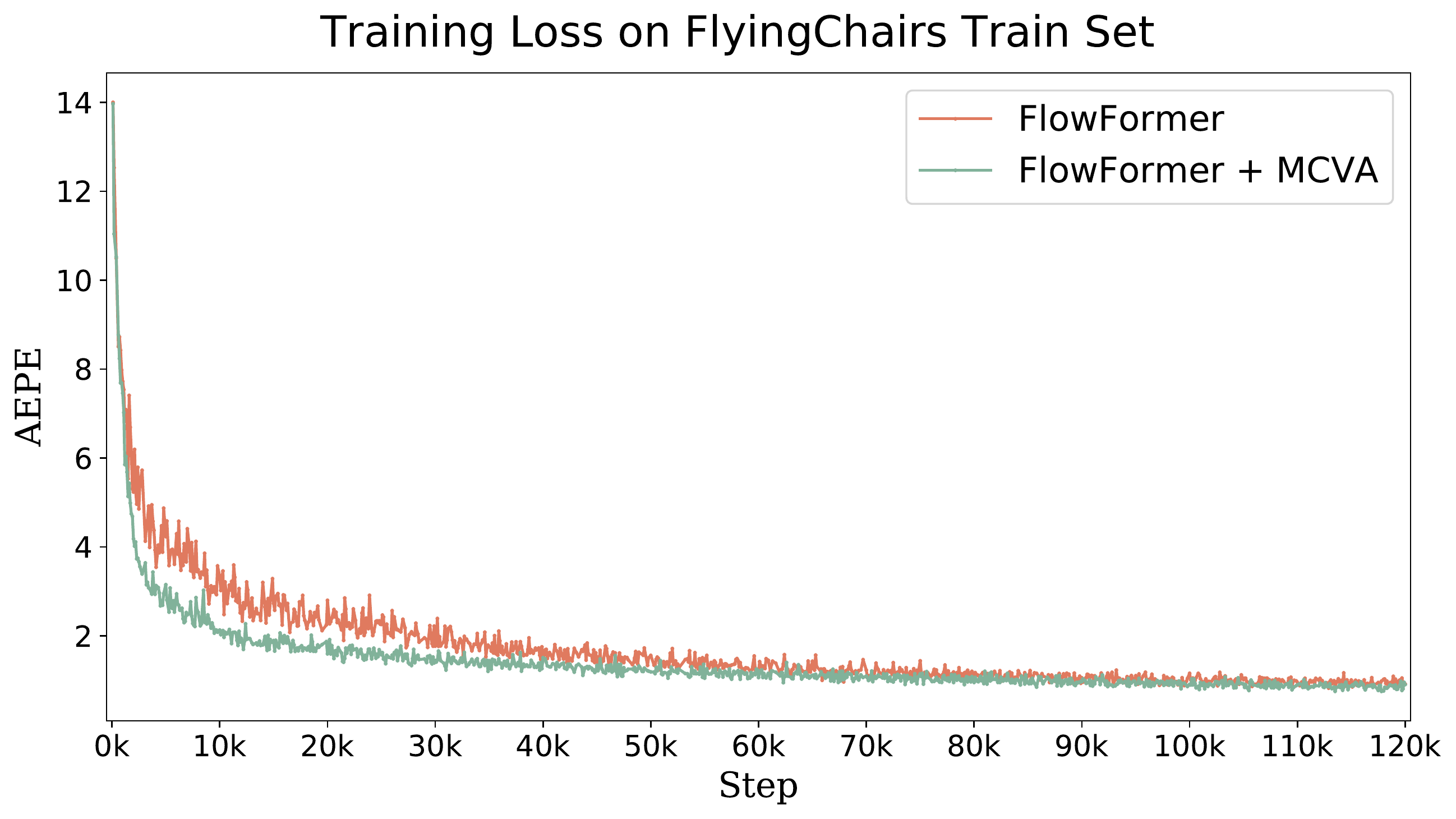}
        \includegraphics[width=\linewidth, trim={0mm 0mm 0mm 0mm}, clip]{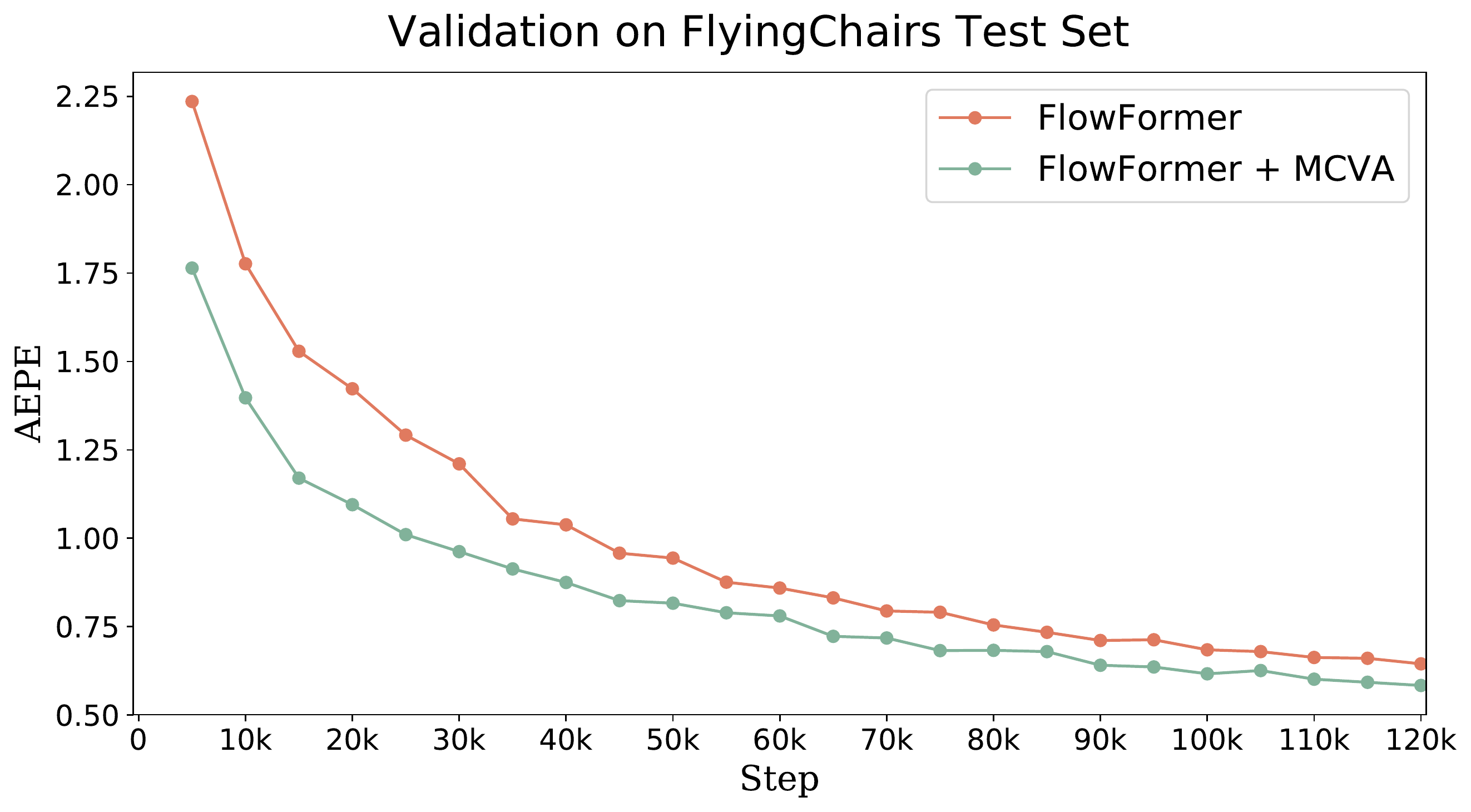}
    }
    \caption{$\textbf{AEPE vs. training-step curves of FlowFormer and FlowFormer+MCVA.}$ FlowFormer+MCVA converges faster and achieves lower validation error.
    }
    \label{Fig: training curve}
\end{figure*}

\noindent \textbf{The Depth of the Reconstruction Head.} We use a light weight MLP as the reconstruction head. 
As shown in Tab.~\ref{Tab: mlp-depth}, the best performance is achieved when the depth is 3, so we use a 3-layer MLP as the reconstruction head when pretraining FlowFormer with MCVA.
Using a shallow single layer as the reconstruction head makes the reconstruction too challenging and leads to inferior performance.
A 5-layer MLP also shows to worse performance because The effective knowledge may be shifted from the AGT to the reconstruction head while the reconstruction head will be dropped in finetuning.

\noindent \textbf{Masking Strategy.}
Masking strategy is one important design of our MCVA. As shown in Tab.~\ref{Tab: masking.}, pretraining FlowFormer with random masking already improves the performance on three of the four metrics. But the proposed block-sharing masking strategy brings even larger gain, which demonstrates the effectiveness of this design. Besides, we observe higher pretraining loss with block-sharing masking than that with random masking, validating that the block-sharing masking makes the pretraining task harder.

\noindent \textbf{Masking Ratio.}
Masking ratio influences the difficulty of the pre-text reconstruction task. We empirically find that the mask ratio of 50\% yields the best overall performance (Tab.~\ref{Tab: masking.}).

\noindent \textbf{Pre-text Reconstruction Design.} The conventional MAE methods aim to reconstruct input data at fixed locations and use the positional encodings as query features to absorb information for reconstruction (the first row of Tab.~\ref{Tab: Pre-text reconstruction design.}). To keep consistent with the dynamic positional query of FlowFormer architecture, we propose to reconstruct contents at random locations (the second row of Tab.~\ref{Tab: Pre-text reconstruction design.}) and additionally use local patches as query features (the third row of Tab.~\ref{Tab: Pre-text reconstruction design.}). The results validate the necessity of ensuring semantic consistency between pretraining and finetuning.

\noindent \textbf{Size of ${\bf q}_{\bf x}^L$.}
During MCVA pertaining, FlowFormer reconstructs a larger cost-map patch ${\bf q}_{\bf x}^L$ with features retrieved from the cost memory. Increasing the size of the ${\bf q}_{\bf x}^L$ will increase the reconstruction difficulty. We show that $15\times 15$ is a good choice for MCVA.

\noindent \textbf{Freezing Image and Context Encoders.}
The FlowFormer architecture has an image encoder to encode visual appearance features for constructing the cost volume, and a context encoder to encode context features for flow prediction. As shown in Tab.~\ref{Tab: freezing encoders.}, freezing the image encoder is necessary, otherwise the model diverges. We hypothesize that the frozen image encoder ensures the reconstruction targets (\textit{i.e.}, raw cost values) to keep static. Freezing the context encoder leads to better overall performance.

\noindent \textbf{Comparisons with Unsupervised Methods.}
We also use conventional unsupervised methods to pretrain FlowFormer with photometric loss and smooth loss following~\cite{stone2021smurf,9477059,Luo_2021_CVPR} and then finetune it in the `C+T' setting as MVCA-FlowFormer. As shown in Tab.~\ref{Tab: unsupervised}, our MCVA outperforms the unsupervised counterpart for pretraining FlowFormer.

\noindent \textbf{FlowFormer with v.s. without MCVA on FlyingChairs.
}
We show the training and validating loss of the training process on FlyingChairs~\cite{dosovitskiy2015flownet} in Fig.~\ref{Fig: training curve}.
FlowFormer+MCVA presents faster convergence during training and better validation loss at the end, which reveals that FlowFormer learns effective feature relationships during MCVA pretraining and benefits the supervised finetuning.

\section{Conclusion}
In this paper, we propose an optical Flow Transformer~(FlowFormer) and Masked Cost Volume Autoencoding (MCVA) to enhance the cost-volume encoder of FlowFormer by pertaining.
MCVA-FlowFormer is the first work that introduces the transformers with MAE pretraining paradigm to optical flow and sheds light on improving optical flow estimation with unlabeled data.
We show that the naive adaptation of MAE scheme to cost volume does not work due to the redundant nature of cost volumes and the incurred pretraining-finetuning discrepancy. We tackle these issues with a specially designed block-sharing masking strategy and the novel pre-text reconstruction task. These designs ensure semantic integrity between pretraining and finetuning and encourage the cost-volume to aggregate information in a long range. Experiments demonstrate clear generalization and dataset-specific performance improvements. 
\textbf{Limitations.} FlowFormer is a transformer-based network architecture, which is slightly sensitive to image size due to the positional encoding. Therefore, FlowFormer needs the tile technique for images that have a quite different size compared to training images.

% or


% \appendices

% \section{}

% \ifCLASSOPTIONcompsoc
%   % The Computer Society usually uses the plural form
%   \section*{Acknowledgments}
% \else
%   % regular IEEE prefers the singular form
% \fi


\ifCLASSOPTIONcaptionsoff
  \newpage
\fi

{\small
\bibliographystyle{IEEEtran}
\bibliography{ref}
}

\ifCLASSOPTIONcompsoc
  \section*{Acknowledgments}
\else
  \section*{Acknowledgment}
\fi

This work is supported in part by Centre for Perceptual and Interactive Intelligence Limited, in part by the General Research Fund through the Research Grants Council of Hong Kong No. 14204021.

% We would like to thank Eric Ryan Chan for sharing the well structured code of the wonderful project EG3D. We thank Yu Deng for providing the evaluation code of the Disentanglement Score, and Jianzhu Guo for providing the pre-trained face reconstruction model. We are also indebted to thank Xingang Pan, Han Zhou, KwanYee Lin, Jingtan Piao, and Hang Zhou for their insightful discussions.

\ifCLASSOPTIONcaptionsoff
  \newpage
\fi

\newpage

\begin{IEEEbiography}[{\includegraphics[width=1in,height=1.25in,clip,keepaspectratio]{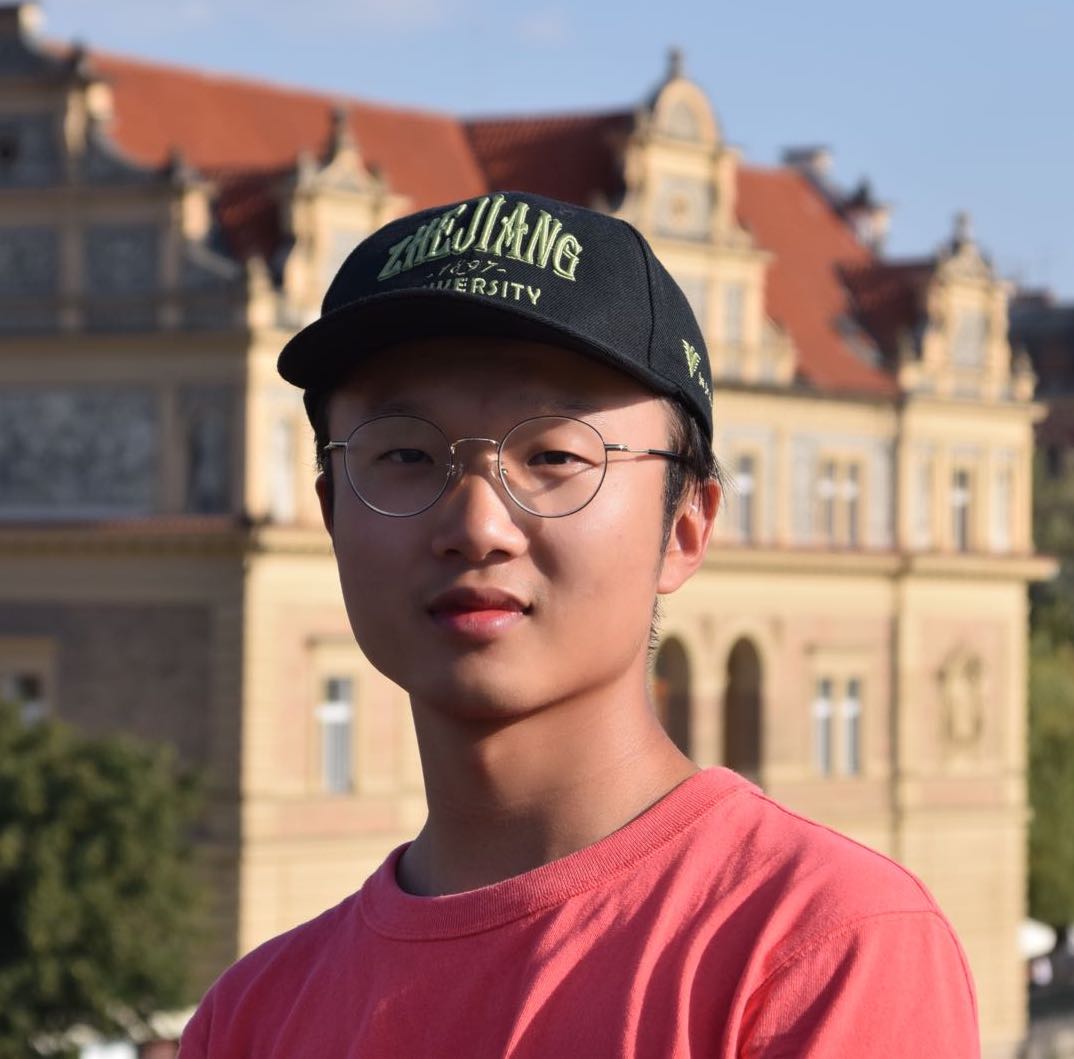}}]{Zhaoyang Huang}
Zhaoyang Huang is currently a Ph.D. candidate in electronic engineering from The Chinese University of Hong Kong, advised by Hongsheng Li. 
He received his bachelor's and master’s degree in computer science from Zhejiang University.
His research focuses on image correspondence learning, video editing, and generative neural radiance fields. 
\end{IEEEbiography}

\begin{IEEEbiography}[{\includegraphics[width=1in,height=1.25in,clip,keepaspectratio]{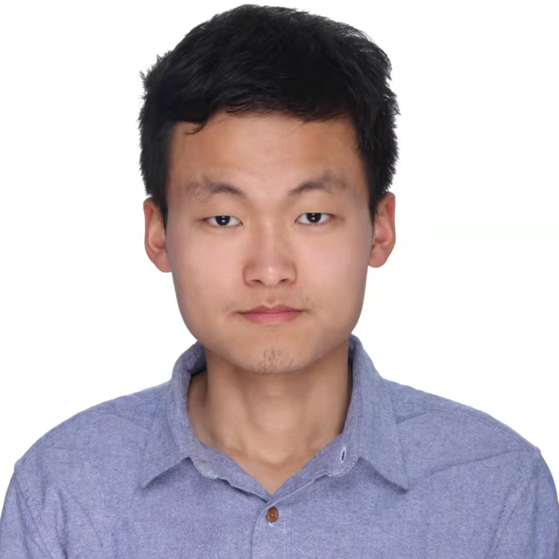}}]{Xiaoyu Shi}
Xiaoyu Shi is currently a Ph.D. candidate in electronic engineering from The Chinese University of Hong Kong, advised by Hongsheng Li. He received his bachelor's degrees in computer science from Zhejiang University and Simon Fraser University. His research focuses on optical flow estimation. 
\end{IEEEbiography}

\begin{IEEEbiography}[{\includegraphics[width=1in,height=1.25in,clip,keepaspectratio]{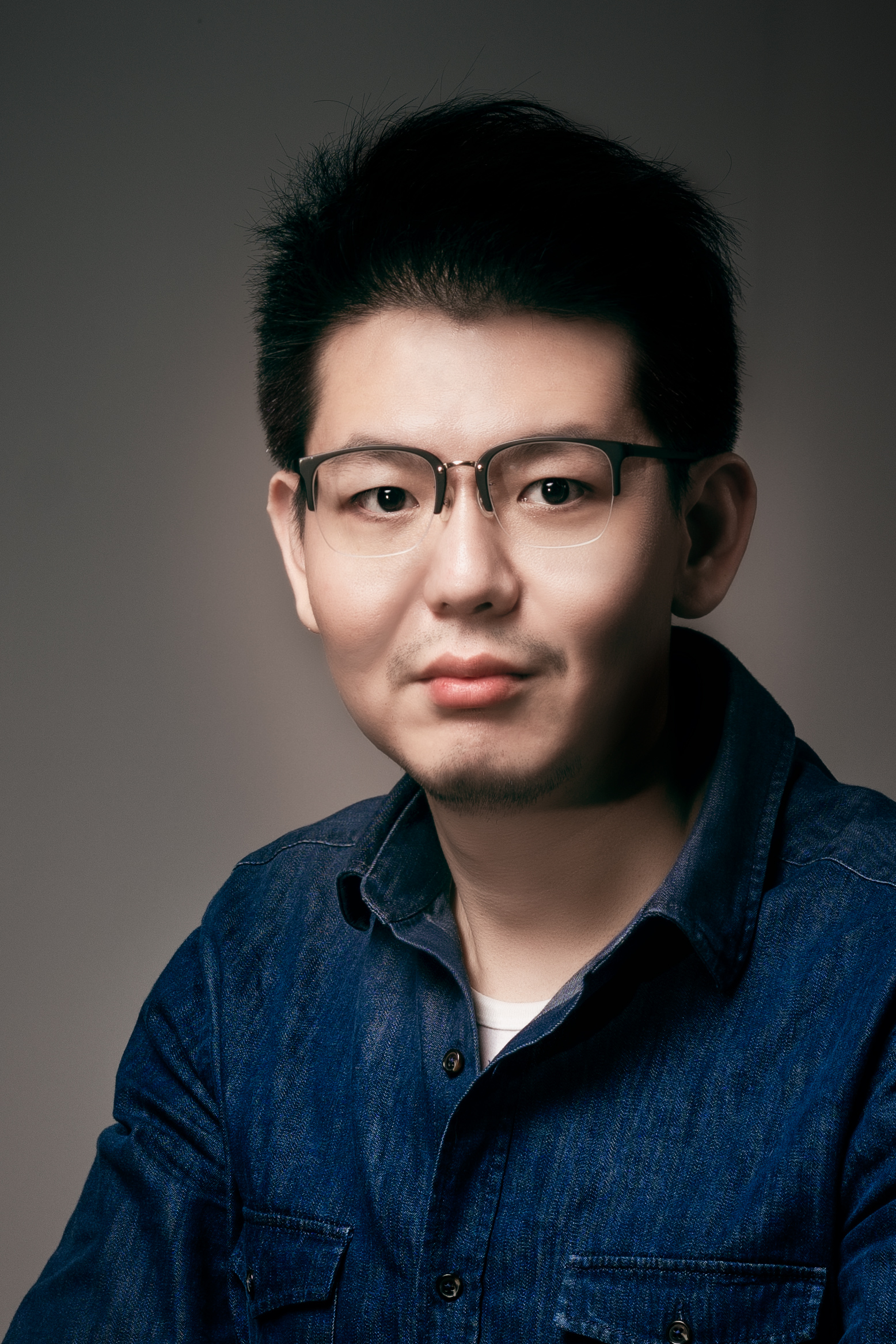}}]{Chao Zhang}{Chao Zhang received the BS degree in Electrical Engineering from the Tianjin University, Tianjin, China, in 2009. And the PhD degree in computer science from The University of New South Wales, NSW, Australia in 2014. He is currently a staff engineer at the SAIT China Lab, Samsung Research China, Beijing. His research interests include computer vision, camera ISP, and machine learning.}
\end{IEEEbiography}

\begin{IEEEbiography}[{\includegraphics[width=1in,height=1.25in,clip,keepaspectratio]{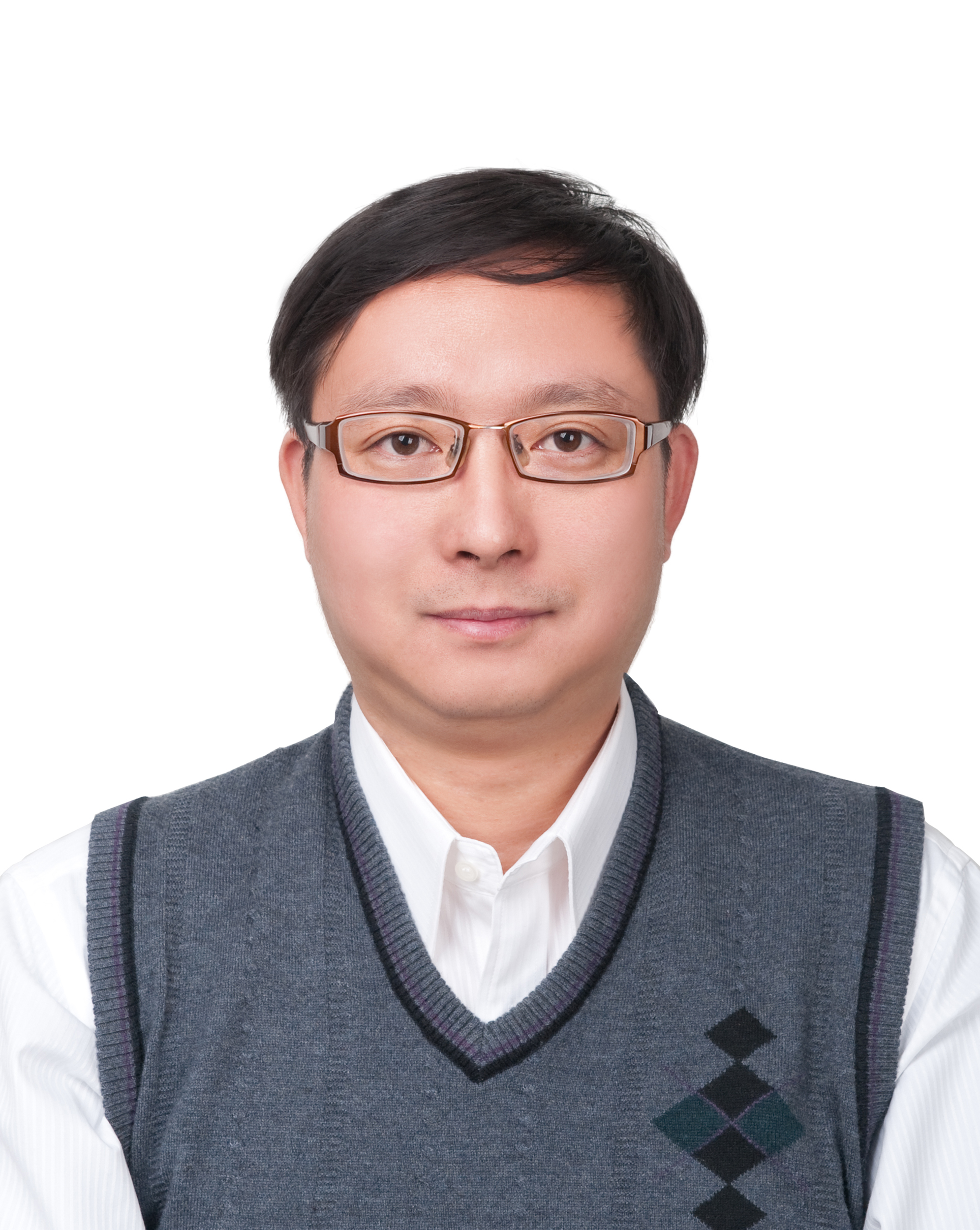}}]{Qiang Wang} Qiang Wang received the BS, MS and Ph.D degrees in computer science from Tsinghua University, Beijing, China, in 1998, 2001 and 2004, respectively. He is currently a principal researcher at Samsung Research China - Beijing, Samsung Electronics. His research interests include computer vision, machine learning, and computer graphics.
\end{IEEEbiography}

\begin{IEEEbiography}[{\includegraphics[width=1in,height=1.25in,clip,keepaspectratio]{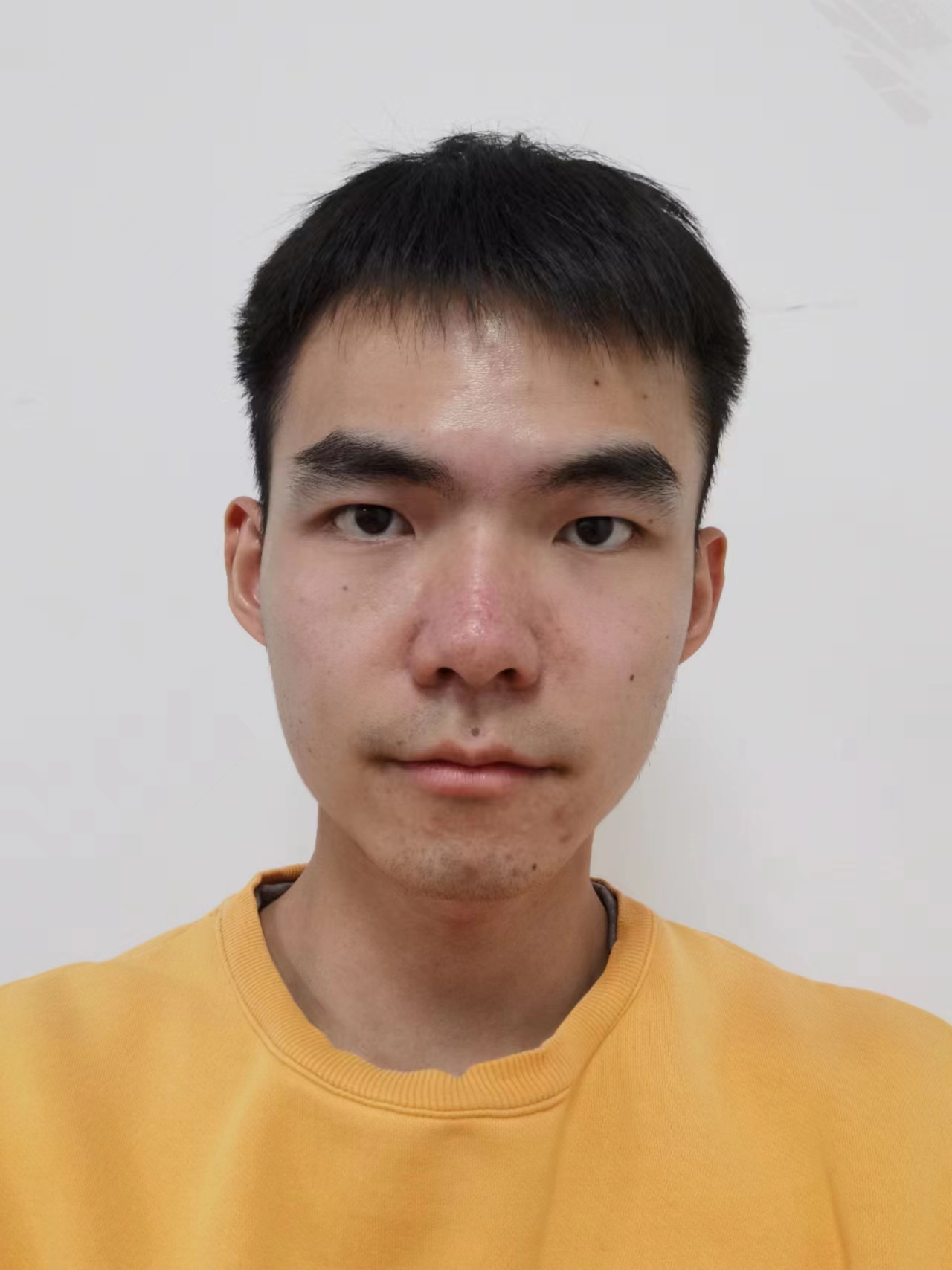}}]{Yijin Li}{Yijin Li is currently a Ph.D. candidate in computer science from Zhejiang University, advised by Guofeng Zhang. 
He received his bachelor's degree in software engineering from Sun Yat-sen University.
His research focuses on event-based vision, 3D reconstruction, and augmented reality.}
\end{IEEEbiography}

\begin{IEEEbiography}[{\includegraphics[width=1in,height=1.25in,clip,keepaspectratio]{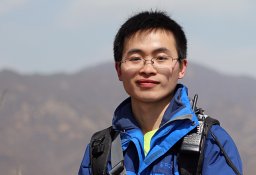}}]{Hongwei Qin} Hongwei Qin leads a team that focuses on AI ISP and AI Video Codec (including software and silicon) for next
generation AI-Camera. Before that, he got his PhD and Bachelor of Engineering from Tsinghua
University in Jun. 2017 and Jun. 2012 respectively. 
He works in SenseTime from 2015 till now.
\end{IEEEbiography}

\begin{IEEEbiography}[{\includegraphics[width=1in,height=1.25in,clip,keepaspectratio]{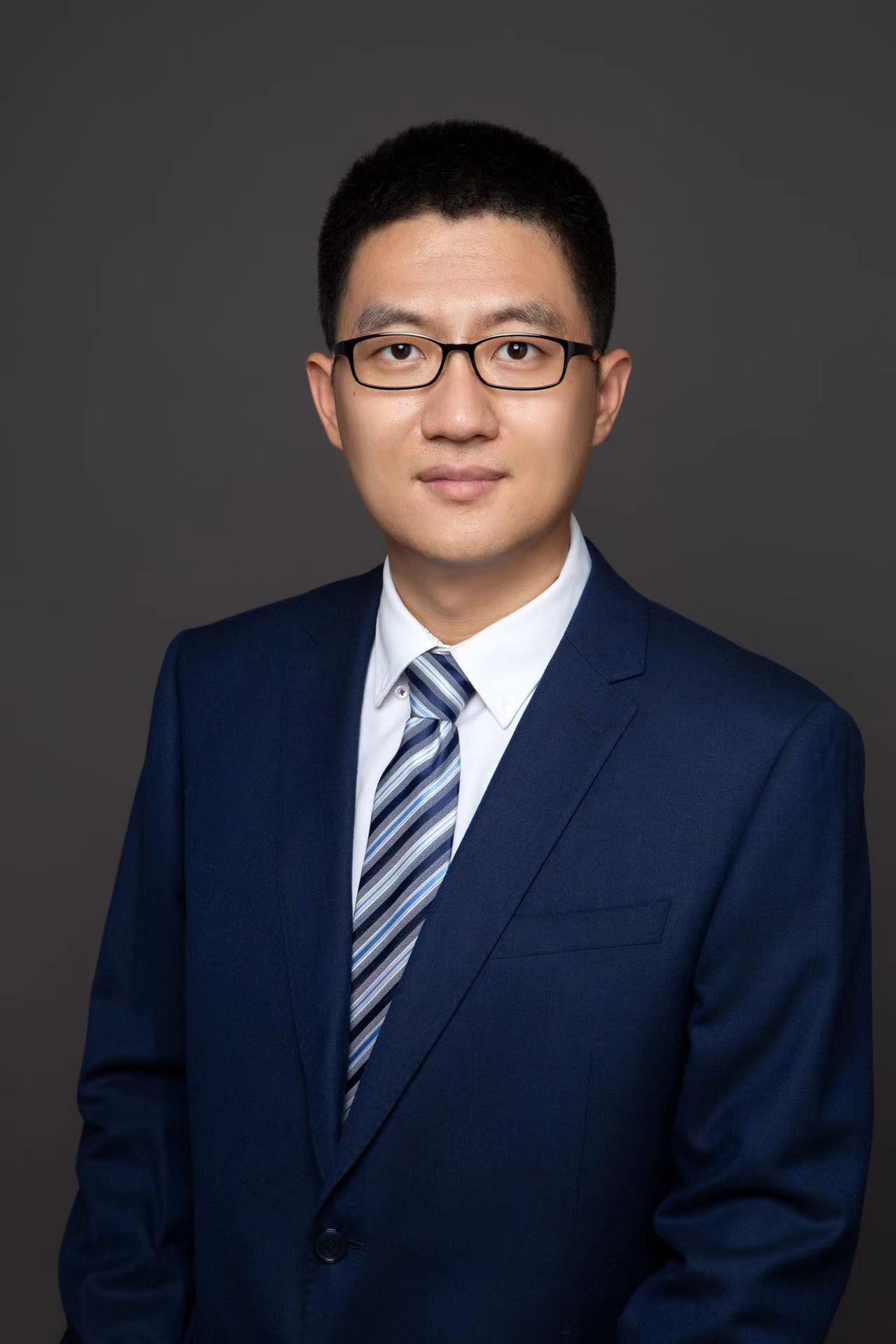}}]{Jifeng Dai}
Jifeng Dai is an Associate Professor at the Department of Electronic Engineering of Tsinghua University. His current research focus is on deep learning for high-level vision.
Prior to that, He was an Executive Research Director at SenseTime Research, headed by Professor Xiaogang Wang, between 2019 and 2022. He was a Principle Research Manager in Visual Computing Group at Microsoft Research Asia (MSRA) between 2014 and 2019, headed by Dr. Jian Sun.
He got his Ph.D. degree from the Department of Automation, Tsinghua University in 2014, under the supervision of Professor Jie Zhou. 
\end{IEEEbiography}

\begin{IEEEbiography}[{\includegraphics[width=1in,height=1.25in,clip,keepaspectratio]{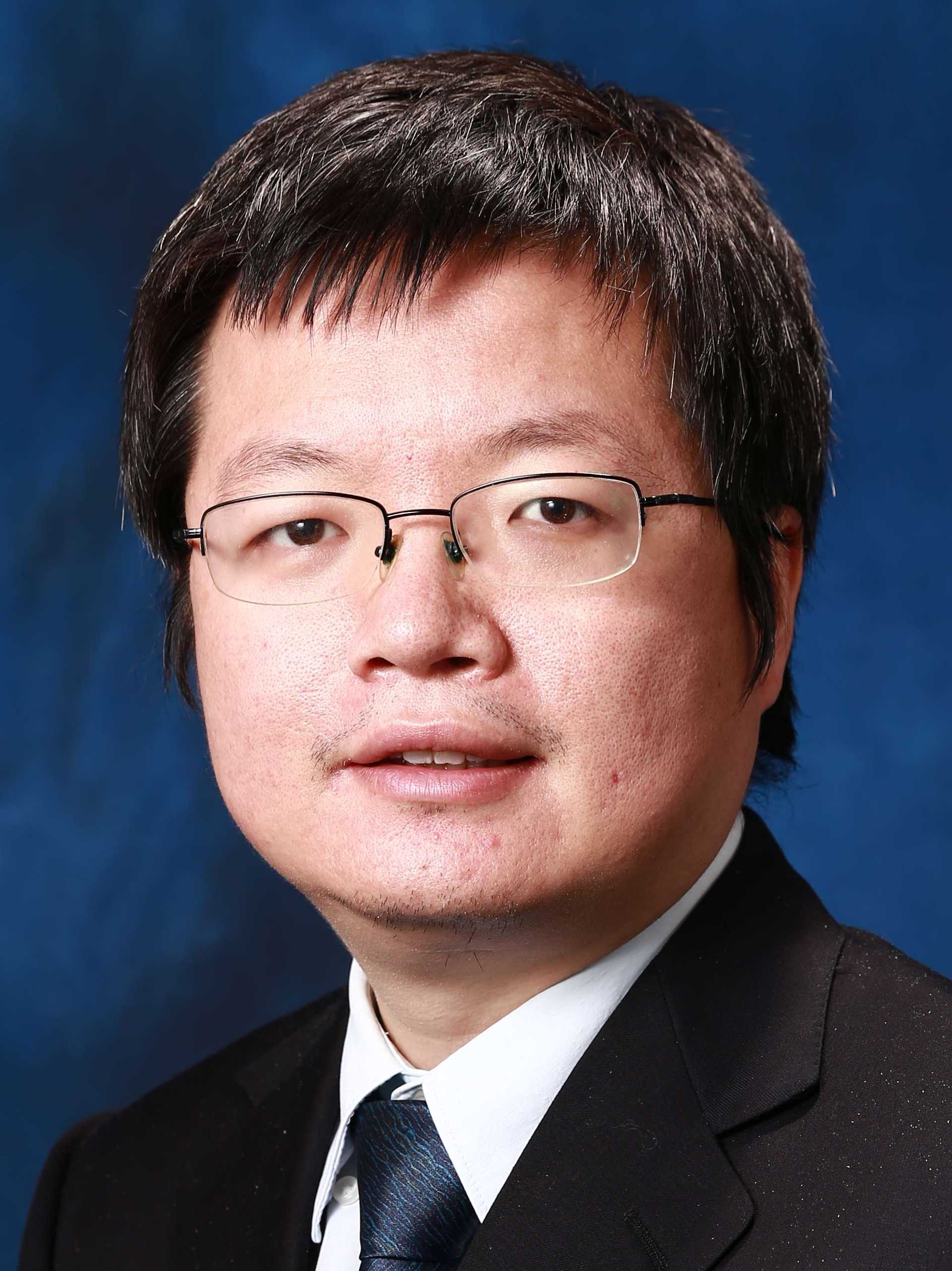}}]{Xiaogang Wang}
Xiaogang Wang received his Bachelor degree in Electrical Engineering and Information Science from the Special Class of Gifted Young at the University of Science and Technology of China, MPhil. degree in Information Engineering from the Chinese University of Hong Kong, and Ph.D. degree in Computer Science from Massachusetts Institute of Technology. He is an assistant professor in the Department of Electronic Engineering at the Chinese University of Hong Kong since August 2009. He was the Area Chair of IEEE International Conference on Computer Vision (ICCV) 2011. He received the Outstanding Young Researcher in Automatic Human Behaviour Analysis award in 2011, and Hong Kong Early Career Award in 2012.
\end{IEEEbiography}

\begin{IEEEbiography}[{\includegraphics[width=1in,height=1.25in,clip,keepaspectratio]{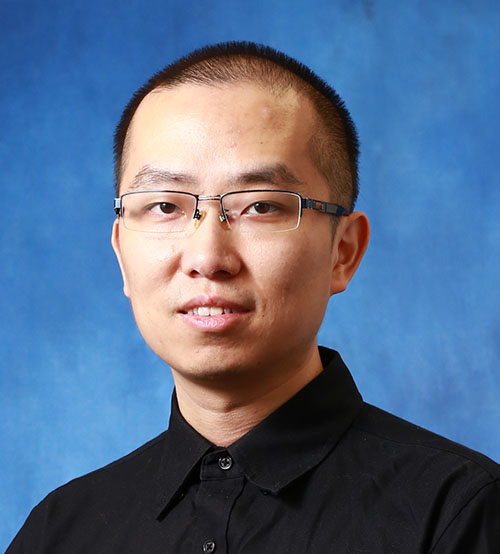}}]{Hongsheng Li}
Hongsheng Li (Member, IEEE) received the BS degree in automation from the East China University of Science and Technology, Shanghai, China, in 2006, and the MS and PhD degrees in computer science from Lehigh University, Bethlehem, Pennsylvania, in 2010 and 2012, respectively. He is currently an associate professor at the Department of Electronic Engineering, The Chinese University of Hong Kong. His research interests include computer vision, medical image analysis, and machine learning.
\end{IEEEbiography}

\end{document}